\documentclass[10pt,twocolumn,letterpaper]{article}

\usepackage[pagenumbers]{cvpr} % To force page numbers, e.g. for an arXiv version
\usepackage{graphicx}
\usepackage{subcaption}
\usepackage{subcaption}
\usepackage{booktabs}
\usepackage{multirow}
\usepackage{bm}
\usepackage{amsmath}
\usepackage{amsfonts}
\usepackage{amssymb}
\usepackage{wrapfig}
\usepackage{url}
\usepackage{colortbl}
\usepackage{subfiles} 

\usepackage[dvipsnames]{xcolor}

\definecolor{cvprblue}{rgb}{0.21,0.49,0.74}
\usepackage[pagebackref,breaklinks,colorlinks,citecolor=cvprblue]{hyperref}

\def\paperID{2194} % *** Enter the Paper ID here
\def\confName{CVPR}
\def\confYear{2025}

\title{Learning Extremely High Density Crowds as Active Matters}

\author{Feixiang He$^2$, Jiangbei Yue$^3$, Jialin Zhu$^2$, Armin Seyfried$^4$, Dan Casas$^5$\thanks{work done prior to joining Amazon}, Julien Pettré$^6$, He Wang $^{1,2}$\thanks{Corresponding author, he\_wang@ucl.ac.uk}\\$^1$AI Centre, University College London, UK\ \ \ \ $^2$University College London, UK\\ $^3$University of Leeds, UK\ \ \ \ $^4$ Forschungszentrum J\"{u}lich, Germany \\$^5$Universidad Rey Juan Carlos, Spain\ \ \ $^6$ INRIA Rennes, France}

\newcommand{\vect}[1]{{{\bf{#1}}}}
\newcommand{\mat}[1]{{{\bf{#1}}}}
\newcommand{\func}[1]{{{\bm{#1}}}}

\begin{document}
\maketitle
\begin{abstract}
Video-based high-density crowd analysis and prediction has been a long-standing topic in computer vision. It is notoriously difficult due to, but not limited to, the lack of high-quality data and complex crowd dynamics. Consequently, it has been relatively under studied. In this paper, we propose a new approach that aims to learn from in-the-wild videos, often with low quality where it is difficult to track individuals or count heads. The key novelty is a new physics prior to model crowd dynamics. We model high-density crowds as active matter, a continumm with active particles subject to stochastic forces, named `crowd material'. Our physics model is combined with neural networks, resulting in a neural stochastic differential equation system which can mimic the complex crowd dynamics. Due to the lack of similar research, we adapt a range of existing methods which are close to ours for comparison. Through exhaustive evaluation, we show our model outperforms existing methods in analyzing and forecasting extremely high-density crowds. Furthermore, since our model is a continuous-time physics model, it can be used for simulation and analysis, providing strong interpretability. This is categorically different from most deep learning methods, which are discrete-time models and black-boxes.
\end{abstract}

\section{Introduction}
\label{sec:intro}

Video-based crowd forecasting/analysis has been a long-standing problem in computer vision \cite{junior2010crowd}, and is especially crucial in public safety and event organization ~\cite{he2013review}. In this area, one of the most difficult scenarios is extremely high-density crowds. We are particularly interested in crowds where the density is at least 5$p/m^2$ (people per square meter). At this density level, forecast/analysis becomes notoriously challenging as the dynamics can become violently chaotic \cite{golas2014continuum}, resulting in life-threatening crushes~\cite{crowd_risky}, which is under-studied in computer vision.

Early research has primarily focused on relatively low density, with empirical modeling ~\cite{ali2007lagrangian,narain2009aggregate,helbing1995social,van2008reciprocal,treuille2006continuum} and data-driven methods ~\cite{karamouzas2018crowd, lee2007group, lopez2019character,gupta2018social,alahi2016social,xia2022cscnet}. While the former is explicit and explainable but inaccurate in prediction, the latter is accurate in prediction but lacks of explainability ~\cite{wu2017crowd, van2021algorithms, wolinski2014parameter}. A recent trend is hybrid models, combining black-box neural networks (NNs) with explicit models, leveraging the advantages of both methodologies. These methods combine NNs with differential equations~\cite{Jiang_trajectory_2022}, or with physics conditioned on social interactions~\cite{chen2024social, xiang2024socialcvae}, or with physics and symbolic regression \cite{mo2024pi}. However, these approaches are designed mainly for low-density crowds, where the microscopic motion is simple and trajectory data can be reliably obtained, which is unlikely in high-density crowds from in-the-wild videos. 
Furthermore, merely focused on local interactions, they lack the ability to capture large scale/global flow patterns~\cite{van2021sph}.A very recent attempt introduced a hydrodynamics-informed neural network for simulating dense
crowds~\cite{zhou2024hydrodynamics} utilizing optical flows. However,  it fails to account for the self-propelled characteristics of crowds and is not suitable for agent-based simulation.

Extremely high-density crowds present unique challenges for prediction and analysis. The foremost is the scarcity of high-quality data. For high-density crowds, often the only available data is Closed-Circuit Television (CCTV) videos, because cameras are the \textit{only} sensors allowed in many public spaces due to their non-invasiveness. As a result, it is imperative to be able to work directly with these videos. However, in public spaces, the cameras are often far from the crowd and the data is noisy therefore unlikely to provide accurate individual motion, if at all, even with crowd counting/localization methods \cite{liu2023point,liu2024consistency,liang2022end,abousamra2021localization}. This makes agent-based analysis ~\cite{Jiang_trajectory_2022,yue2023human,he2020informative,wang_path_2016,wang2016globally,wang_trending_2016}, which relies on trajectory estimation, impractical. Furthermore, complex dynamics arise when crowd density exceeds 5$p/m^2$. Vastly different from low-density crowds, high-density crowds behave like an active matter \cite{pincce2016disorder}, \eg a continumm with particles moving non-deterministically caused by autonomous behaviors and self-propulsion even when being physically constrained by the surroundings. Consequently, the local interactions become significantly more pronounced, sometimes causing spatiotemporal fluctuations similar to waves~\cite{helbing2007dynamics,adrian2020crowds} leading to life-threatening states in crowds~\cite{sieben2023inside}. Accurately modeling such dynamics remains an open problem. In the past, there has been parallel drawn between high-density crowds and continuum~\cite{narain2009aggregate}. Such parallels lead to explicit models but they are not learnable, \ie hard to be adapted to specific crowds.

To address the aforementioned challenges, we propose a new learnable active matter model for high-density crowds.  Since crowds show distinctive dynamics compared with common homogeneous materials \eg water, we design a new material model, named ``crowd material'', which has a special stress-strain relation to reflect the empirical observations on individuals in high-density crowds. Additionally, to mimic active matters, this new material model is augmented by integrating the collective dynamics of self-propelled agents, based on the Toner-Tu (TT) equation \cite{toner1995long}, enabling us to model and learn random individual motions. Next, our model is described by a new neural stochastic differential equation system.  To directly learn from raw videos, we treat the optical flows from videos as noisy observations of the underlying velocity field of the continuum. Meanwhile, we also model individuals as active 2D particles. Note that treating people as 2D particles in a continuum is a common practice in macroscopic high-density crowd research~\cite{jiang2010continuum}. This dual continuum-particle perspective naturally leads us to a new Material Point Method (MPM)~\cite{jiang2016material}, where the Eulerian grid discretizes the space, and the Lagrangian particles represent the individuals. We refer to this new MPM framework as CrowdMPM. Finally, we utilize NNs to learn the active forces and key material parameters.

Our model is designed to handle optical flows estimated from real-world videos. In addition, due to the scarcity of high-density crowd data and our ambition to learn from in the wild data, we use data from lab environments as well as public spaces. These datasets are all high-density scenarios with diverse dynamics such as evacuation, music festivals, \etc. Next, due to the lack of similar research, we adapt both physics-based and data-driven methods for comparison. The experiments show that our model outperforms the alternatives in capturing, analyzing and predicting high-density crowd dynamics. Our contributions include:
\begin{itemize}
    \item a new video-based analysis and prediction framework for extremely high-density crowds.
    \item a new differentiable physics model for high-density crowd dynamics.
    \item a new crowd material model capable of capturing crowds as active matters.
    \item a new Eulerian-Lagrangian scheme, CrowdMPM.
\end{itemize}

\section{Related Work}
\label{sec:related_work}
\subsection{Behavior Understanding from Videos}

Traditional approaches for crowd behavior analysis are individual-based~\cite{cai2006robust,wang_path_2016,wang_trending_2016,wang2016globally,He_Informative_2020} and global-based~\cite{hu2008learning, ali2008floor, yang2009video}. The former treats a crowd as a collection of individuals and analyzes behaviors by segmenting or detecting individuals; conversely, the latter treats the crowd as a single entity without segmentation. In dense scenes, it is very difficult to employ individual-based approaches as tracking individuals is extremely challenging. Benefited from the data-fitting ability of deep learning (DL), DL-based approaches are proposed to understand human behaviors in crowded scenarios, such as behavior classification~\cite{wu2017crowd}, anomaly detection~\cite{singh2020crowd} or predicting pedestrian trajectories~\cite{gupta2018social, alahi2016social}. However, they are not designed for predicting high-density crowd dynamics directly from optical flow and cannot be used for simulation with changed environments and densities. 

\subsection{Differentiable Physics}
Solving differentiable equations (DE) with deep learning has
recently spiked strong interests. The research can be grouped into different categories according to how deep learning is involved. NNs can be used to generate finite element meshes~\cite{Zhang_MeshingNet_2020,zhang_meshingnet3d_2021}. Neural differential equation models~\cite{shen2021high, Jiang_trajectory_2022,yue2023human} attempt to parameterize part of the DEs by neural networks. Differentiable physics~\cite{gong2022fine, liang2019differentiable, werling2021fast} focus on making the whole simulation process differentiable. Physics-informed neural networks~\cite{raissi2019physics,song2024loss,zhou2024hydrodynamics} aim to bypass the DE solve using NNs for prediction. Highly inspired by the research above, we propose a novel differentiable physics model for forecasting high-density crowds.

\subsection{Video Prediction}
Recently, video prediction has become a prominent area of research, focusing on forecasting future frames based on past observations, which aligns closely with our objective of predicting future optical flow frames. Most existing video prediction methods like  recurrent neural networks ~\cite{wang2022predrnn, wang2018predrnn++}, convolutional neural networks~\cite{gao2022simvp, zhong2023mmvp}, or attention mechanisms~\cite{tan2023temporal}  are tailored for RGB inputs rather than optical flow. Shifting the input from RGB frames to optical flow enables us to focus directly on motion dynamics, a critical factor in crowd movement prediction. However, these methods typically require a long input sequence and rely on deep learning architectures with large numbers of parameters, which limits interpretability. In contrast, our model requires only a single optical flow frame as input and maintains interpretability, providing a more efficient and explainable approach to predicting crowd dynamics.

\section{Methodology}
\label{sec:methodology}
\begin{figure*}[tb]
    \centering
    \includegraphics[width=0.98\textwidth]{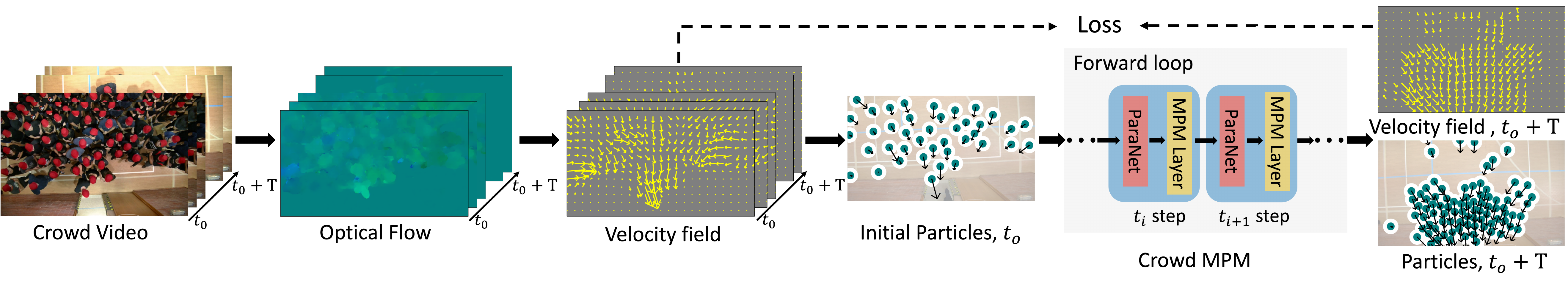}
    \vspace{-0.3cm}
    \caption{Overview. From left to right: optical flow estimation, velocity field generation, initial particle sampling, crowd simulation and loss calculation.}
    \label{fig:framework}
    \vspace{-0.3cm}
\end{figure*} 
Our whole framework is shown in \cref{fig:framework}. Our model takes optical flows as input from videos, simulate the crowd movements, then learn the model parameters based on the difference between the predicted and the observed crowd motions. For simplicity, we only give the key equations below and refer the readers to the supplementary materials (SMs) for details.

\subsection{Crowd as an Active Matter}
High-density crowds bear strong similarity with continuum ~\cite{treuille2006continuum}. We start by modeling the continuum with mass and momentum conservation:
\begin{align}
    \frac{D\rho}{Dt} + \rho\nabla\cdot\vect{v} = 0 \,\, \text{and} \,\, \rho\frac{D\vect{v}}{Dt} = \nabla\cdot\vect{\sigma^{cm}} + \rho\vect{b} + \vect{f^{act}}
\label{eq:governing}
\end{align}
where $\rho$ is density and $\frac{D}{Dt}$ is the material derivative~\cite{temam2001navier}. $\vect{v}$ is the velocity field. $\rho\vect{b}$ denotes the body force (\eg gravity). The main difference between \cref{eq:governing} and a standard formulation is the two terms, the new crowd material stress $\vect{\sigma^{cm}}$, and the new stochastic active force $\vect{f^{act}}$. The new learnable $\vect{\sigma^{cm}}$ is key to our ``crowd material'' property explained in \cref{sec:learnableStress}, and the new learnable active force $\vect{f^{act}}$ captures the random motions caused by active self-propulsion explained in \cref{sec:learnableActiveForces}. We keep $\rho\vect{b}$ for forces like global attraction, if needed. Overall since we learn the parameters in $\vect{\sigma^{cm}}$ and $\vect{f^{act}}$ by neural networks, \cref{eq:governing} is a neural stochastic differential equation system.

\subsection{CrowdMPM}
\label{sec:crowd_mpm}
To solve \cref{eq:governing}, Eulerian methods and Lagrangian methods are the two main approaches. Eulerian methods approximate quantities (\eg. velocity) on a spatial grid over a continuous domain, while Lagrangian methods discretize the continuum into mass particles and trace the quantities on them. However, the challenge is we only have Eulerian data but need to model Lagrangian behaviors. The only data available is video, where we can estimate the optical flows. The optical flows can be seen as a noisy observation of the velocity field of the underlying continuum, \ie Eulerian data. Meanwhile, we need to model active forces on individuals, \ie Lagrangian behaviors. Eulerian methods are limited in modeling the individual behaviors explained later which is crucial for active forces, and Lagrangian methods do not work well with our Eulerian data as they can only  compute quantities in areas when there are particles which cannot guarantee to cover the whole space. 

Therefore, we choose material point methods (MPM)~\cite{hu2018moving}. MPM is a hybrid Eulerian-Lagrangian method, which uses both spatial grids and mass particles. However, in our CrowdMPM, the particles are different from those in standard MPM which only serve as collocation points for integration and not corresponding to any entities. Our particles are individuals. This perspective is seemingly trivial but crucial in modeling the crowd material introduced later in \cref{sec:learnableStress} and \cref{sec:learnableActiveForces}.

Typically, MPM takes three steps to solve an equation: (1) Particle-to-grid transfer (P2G) (2) Grid Operation (GO) and (3) Grid-to-particle transfer (G2P). First, P2G transfers mass, velocity, and pressure from particles to the grid nodes, then GO solves the momentum equation under boundary conditions on the grid nodes, and finally, the updated velocity and deformation are transferred back to particles from the grid nodes. Here, we directly give the final discretized update scheme and leave the details in the SM. 

\begin{figure*}[tb]
    \centering
    \includegraphics[width=0.8\textwidth]{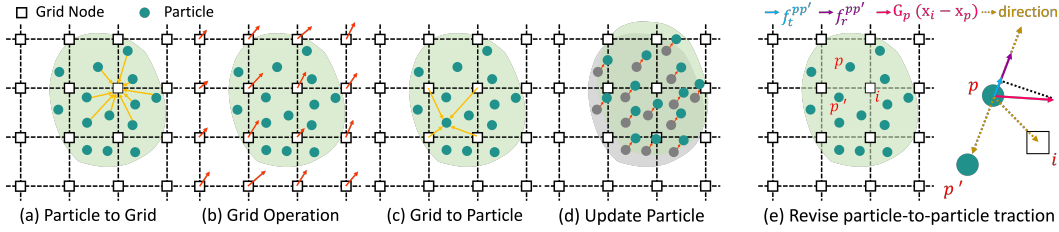}
    \vspace{-0.3cm}
    \caption{(a)-(d) Standard MPM. (e): particle-to-particle traction modeling.}
    \label{fig:mpm}
    \vspace{-0.4cm}
\end{figure*}

\paragraph{Notation} A standard MPM discretization~\cite{hu2018moving} is shown in \cref{fig:mpm}. Formally, we use subscript $i$ to index variables on a 2D grid with cell size $\Delta$x, and $p$ to index variables on particles, and superscript indicating time steps. Node $i$ and particle $p$ have attributes including mass $m_i$, $m_p\in\mathbb{R}$, position $\vect{x}_i$, $\vect{x}_p\in\mathbb{R}^2$, and velocity $\vect{v}_i$, $\vect{v}_p\in\mathbb{R}^2$. Each particle has an initial (undeformed) volume $V_p^0$ at $t=0$. $\mat{C}_p$ is the affine velocity gradient ~\cite{jiang2015affine} and $\mat{P}_p=J_p\sigma^{cm}\mat{F}_p^{-T}$ is the PK1 (the first Piola–Kirchhoff) stress, where $\mat{F}_p$ is the deformation gradient and $J_p$ is the determinant of $\mat{F}_p$ ~\cite{jiang2016material}. $\phi$ is a quadratic B-spline function~\cite{hu2018moving}. We use $\Delta t$ to denote the time interval between time steps. 

At time $n$, we compute the P2G step via:
\begin{align}
&m_i^n = \sum_p w_{ip}^nm_p \nonumber\\
&m_i^n\vect{v}_i^n = \sum_p w_{ip}^n[m_p\vect{v}_p^n + m_p\mat{C}_p^n(\vect{x}_i-\vect{x}_p^n)],
\label{eq:p2g}
\end{align}
where $w_{ip}^n = \phi(\vect{x}_i - \vect{x}_p^n)$. Then, we update velocities in the GO step via:
\begin{equation}
    \vect{v}_i^{n+1} = \vect{v}_i^{n} + \Delta t \vect{f}_i^n/m_i, 
\label{eq:gridUpdate}
\end{equation}
where $\vect{f}_i^n=\vect{f}_i^{n,st} + \vect{f}_i^{n, act} + \vect{f}_i^{n, bd}$ is specified by our new stress and active forces explained later. Boundary conditions are also considered in GO to refine $\vect{v}_i^{n+1}$:
\begin{equation}
    \vect{v}_i^{n+1} = \textbf{BC}(\vect{v}_i^{n+1}) = \vect{v}_i^{n+1} - \gamma\vect{n}\left< \vect{n},\vect{v}_i^{n+1}\right>,
\end{equation}
where $\vect{n}$ is the surface normal of the boundary, $\gamma$ is a scalar. $\textbf{BC}$ is a (dampened) no-slip boundary~\cite{day1990no}. 

Lastly, we update the particle velocity $\vect{v}_p^{n+1}$, velocity gradient $\vect{C}_p^{n+1}$, deformation gradient $\vect{F}_p^{n+1}$ and position $\vect{x}_p^{n+1}$ by gathering information from their neighboring grid nodes, \ie the G2P step:
\begin{align}
    &\vect{v}_p^{n+1} = \sum_i w_{ip}^n\vect{v}_i^{n+1}, \,\, \vect{x}_p^{n+1} = \vect{x}_p^n + \Delta t \vect{v}_p^{n+1},  \\
    &\mat{C}_p^{n+1} = \frac{4}{\Delta x^2}\sum_i w_{ip}^{n} \vect{v}_i^{n+1}(\vect{x}_i-\vect{x}_p^{n})^{T}, \\
    &\mat{F}_p^{n+1} =(\mat{I} + \Delta t \mat{C}_p^{n+1})\mat{F}_p^{n}.
    \label{eq:g2p}
\end{align}
Overall, solving \cref{eq:p2g}-\cref{eq:g2p} solves \cref{eq:governing} in every step.

\subsection{New Learnable Stress $\vect{\sigma^{cm}}$}
\label{sec:learnableStress} 
Empirically, the `crowd material' is significantly different from homogeneous materials like water. Firstly, crowds can easily scatter but not be compressed severely~\cite{narain2009aggregate}, due to people cannot be superpositioned, \ie easily stretched but not as easily compressed, which we refer to as \textit{elastic asymmetry}. Next, during compression, there is a multi-phasic resistance change based on the inter-personal distance~\cite{heigeas2010physically}.  The  resistance is small when the distance is larger than a comfortable threshold, but exponentially increased within the threshold until it reaches incompressibility (\eg people in full contact). We refer to this property as \textit{exponential resistance}. Finally, while resisting compression, people can be more tolerant to relative motions such as passing each other at a close distance. This suggests that `crowd material' has less resistance in shearing and rotations, to which we refer as \textit{compression dominance}.

The afore-mentioned observations dictate that we need to design a new material model that can capture these features and is also learnable. To this end, we propose a new learnable stress inspired by the above observations. First, unlike particles in standard MPM as collocation points and not corresponding to any entities, \textit{we need to model the strain-stress between particles directly, not via the grid nodes as in standard MPM}. So we treat each individual as a Lagrangian particle.  This leads to a new stress $\vect{\sigma^{cm}}$ below for the force $\vect{f}_i^{st}$ in \cref{eq:gridUpdate}.

To model the elastic asymmetry, we choose a Cauchy stress for weakly compressible fluid~\cite{tampubolon2017multi} $\sigma_p = \epsilon(1 - \frac{1}{J_p})\mat{I}$, where $\epsilon$ is the Young's modulus, therefore, the force $\vect{f}_i^{st}$ in \cref{eq:gridUpdate} can be denoted as:
\begin{align}
\vect{f}_i^{st} &= \sum_p \omega_{ip} \mat{G}_p (\vect{x}_i-\vect{x}_p), \label{eq:stress1} \\
\mat{G}_p = -\frac{4}{\Delta x^2} V_p^0 &\sigma_p J_p = -\frac{4}{\Delta x^2} \epsilon V_p^0(J_p-1)\mat{I}, \label{eq:g_p}
\end{align}
where we omit the time superscripts $n$ for simplicity. \cref{eq:stress1} follows the standard MPM, which ensures the overall small compressibility.

Next, to model the exponential resistance, we augment each particle with a volume, shown in \cref{fig:framework} (Initial Particles), which consists of an incompressible volume (green disc) and a comfortable zone (white disc). We assume that when a particle $p'$ is within the comfortable zone of particle $p$, there is a repulsive force based on their distance.  We use a comfort distance $d_c= 2(r_b - r_a)$ where $r_a$ and $r_b$ are the radii of the incompressible volume and the comfortable zone respectively. Then we model the repulsive force as:
\begin{equation}
\resizebox{0.85\hsize}{!}{$
    \func{f}_{r}^{pp'} =
    \begin{cases}
    -k log(d_{pp'})\vect{e}_{p'p} + \func{f}_{t}^{pp'}& \text{if }  0< d_{pp'} < 1 \\
    \func{f}_{t}^{pp'} & \text{otherwise}
    \end{cases}$}
\label{eq:resistance}
\end{equation}
where $d_{pp'}=\frac{||\vect{x}_p - \vect{x}_{p'}|| -2r_a}{d_{c}}$ is the ratio between the current distance and the comfortable distance and $\vect{e}_{p'p} = \frac{\vect{x}_p - \vect{x}_{p'}}{||\vect{x}_p - \vect{x}_{p'}||}$. $k$ is a positive \textit{learnable} parameter. Therefore, $-k log(d_{pp'})\vect{e}_{p'p}$ models the exponential resistance. 

To model the compression dominance,  we separate the inter-particle compression from other stress components related to shear/rotation. We introduce a term $\func{f}_{t}^{pp'}$ in \cref{eq:resistance}. $\func{f}_{t}^{pp'}$ is the traction at $p$ (due to $\vect{G}_p$) projected along the direction from $p'$ to $p$, which can be calculated as $\func{f}_{t}^{pp'} = \left<\mat{G}_p(\vect{x}_i-\vect{x}_p), \vect{e}_{p'p}  \right>\vect{e}_{p'p}$, shown in \cref{fig:mpm} (e). 

Then we incorporate \cref{eq:g_p} and \cref{eq:resistance} into \cref{eq:stress1} and modify $\vect{f}_i^{st}$ for grid update in \cref{eq:gridUpdate} to:
\begin{equation}
    \centering
    \resizebox{0.85\hsize}{!}{$
    \vect{f}_i^{st} = \sum_p w_{ip}\{\mat{G}_p(\vect{x}_i-\vect{x}_p) + \sum_{p'\in N_p}(\func{f}_{r}^{pp'} - \func{f}_{t}^{pp'})\},
    $}
    \label{eq:resistenceUpdate}
\end{equation}
where $N_p$ is the set of the neighboring particles of $p$. In \cref{eq:resistenceUpdate}, we see that when $\func{f}_{r}^{pp'} = \func{f}_{t}^{pp'}$, \cref{eq:resistenceUpdate} is simply \cref{eq:stress1}, \ie the standard MPM; otherwise, the right hand side becomes $\sum_p w_{ip}\{\mat{G}_p(\vect{x}_i-\vect{x}_p) + \sum_{p'\in N_p}-k  log(d_{pp'})\vect{e}_{p'p} \}$. Essentially, we model an exponential residual force which only impacts compression, not shearing or rotation, between particles.

Overall, our model learns parameters $k$ in \cref{eq:resistance} and Young’s modulus $\epsilon$ in \cref{eq:stress1} through neural networks, equivalent to $\vect{\sigma^{cm}}$ in \cref{eq:governing} being learnable. Considering that these parameters vary for each particle at every time step, we estimate them via:
\begin{equation}
k_p=NN_k(\vect{x}_p, \vect{v}_p, N_p), \,\, \epsilon_p=NN_{\epsilon}(\vect{x}_p, \vect{v}_p, N_p).
\label{eq:stressNNs}
\end{equation}
where $N_p$ represents the neighborhood of particle $p$. $NN_k$ and $NN_{\epsilon}$ include multiple Continuous Convolution (CConv) and fully-connected (FC) layers. The detailed architectures are in the SM. 

\subsection{New Learnable Active Forces $\vect{f^{act}}$}
\label{sec:learnableActiveForces}
Besides the new stress, we know that crowds are also influenced by active forces from individuals \cite{ali2008taming}, \eg when one person stiffens the body to restore balance, or move with the neighbors. Overall, these local actions manifest as random active forces from individuals to the whole system, which can be described by the Toner-Tu (TT) equation ~\cite{toner1995long}:
\begin{align}
\label{eq:tt_momentum}
\frac{\partial \vect{v}}{\partial t} + \lambda(\vect{v}\cdot\nabla)\vect{v} &=  \alpha\vect{v} - \beta|\vect{v}|^2\vect{v}-\nabla P + D_L\nabla(\nabla\cdot\vect{v}) \nonumber\\
& +D_1\nabla^2\vect{v} + D_2(\vect{v}\cdot\nabla)^2\vect{v} + \Tilde{\vect{f}}, 
\end{align}
where $\beta$, $D_1$, $D_2$, $D_L$ are all positive, and $\alpha<0$ in disordered phases and $\alpha>0$ in ordered state. Specifically, $\alpha\vect{v}$ and $-\beta|\vect{v}|^2\vect{v}$ represent self-driven motions and velocity saturation, essential for modeling the collective behavior in high-density active systems. The diffusion terms $D_L\nabla(\nabla\cdot\vect{v})$, $D_1\nabla^2\vect{v}$ and $D_2(\vect{v}\cdot\nabla)^2\vect{v}$ handle the spread of velocity perturbations, while $\Tilde{\vect{f}}$ is a Gaussian random force. 

However, we do not directly use the the TT equation for the active force $\vect{f}^{act}$ in \cref{eq:governing}, because we need to learn the free parameters ($\alpha$, $\beta$, $D_1$, $D_2$, $D_L$). Therefore, we first remove $\nabla P$ as it is caused by the Cauchy stress and is learned in \cref{sec:learnableStress}. Then we separate the right hand side of TT into two parts, $\alpha\vect{v}$ and the rest. This is because $\alpha\vect{v}$ dominates motion alignment caused by active forces. We learn $\alpha$ via $\alpha=NN_{\alpha}(\vect{v})$,
where the neural network $NN_{\alpha}$ is detailed in the SM. 

To learn the rest, we observe that although $\Tilde{\vect{f}}$ is a Gaussian random force, the overall distribution of the rest, \ie $- \beta|\vect{v}|^2\vect{v} + D_L\nabla(\nabla\cdot\vect{v}) +D_1\nabla^2\vect{v} + D_2(\vect{v}\cdot\nabla)^2\vect{v} + \Tilde{\vect{f}}$ is not Gaussian. Therefore, we assume its distribution is a Gaussian in a latent space, captured by a latent variable $z\sim Normal$, so that:
\begin{equation}
    \vect{f}^{act} = NN_{act}(|\vect{v}|^2\vect{v}, \nabla(\nabla\cdot\vect{v}), \nabla^2\vect{v}, (\vect{v}\cdot\nabla)^2\vect{v}), z)
\label{eq:activeForce}
\end{equation}
We formulate $NN_{act}$ as the decoder of a Conditional Variational Autoencoder which is detailed in the SM.

\subsection{Body Forces}
Other than the previous forces, there can be global forces that resembles body forces. For instance, the crowd can be attracted to a goal, or circulate around an area, which can be formulated as a global force that is evenly applied to every unit in the continumm. This is modeled by the $f_p= \rho\vect{b}$ in \cref{eq:governing}. For example, for a goal attraction force~\cite{helbing1995social}, we use $\vect{f}^{gl}_{p} = \frac{1}{\tau}(\vect{v}_{d} \vec e_{pg} - \vect{v}_p)$ where $ \frac{1}{\tau} = \frac{m_p}{\Delta t}$ and $\vect{e}_{pg}=\frac{\vect{x}_g-\vect{x}_p}{||\vect{x}_g-\vect{x}_p||}$ where $\vect{x}_g$ is the goal and $\vect{v}_{d}$ is the preferred speed. For circular motions, we can define a centripetal force: $\vect{f}^{cen}_{p} = m_p \frac{\vect{v}_p^2}{r}$ where $r$ is the circle radius. 

\subsection{Training}
\label{sec:training}
Overall, the learnable parameters are the Young's modulus $\epsilon$ and $k$ in \cref{eq:stressNNs}, the $NN_{\alpha}$, and the $NN_{act}$ in \cref{eq:activeForce}. Owning to the full differentiability of our model, we learn the model via Adam. More details are in the SM.

\begin{figure}[t]
    \centering
    \includegraphics[width=0.48\textwidth]{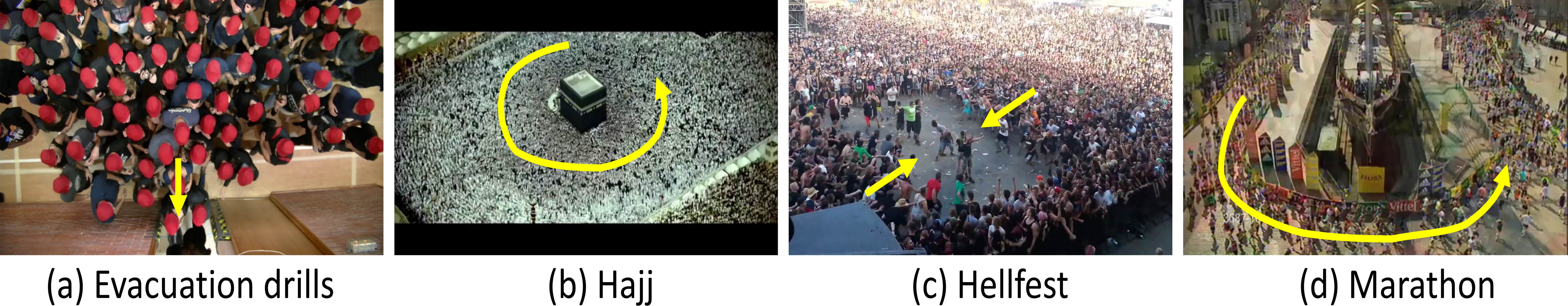}
    \caption{Four crowd scenes with diverse dynamics. The yellow arrows indicate the main movement.}
    \label{fig:dataset}
    \vspace{-0.6mm}
\end{figure}

\section{Experiments}
\label{sec:experiments}
\paragraph{Datasets} We employ Evacuation drill~\cite{garcimartin2016flow}, Hajj~\cite{hajj}, Hellfest~\cite{hellfest} and Marathon ~\cite{ali2007lagrangian} (\cref{fig:dataset}). Evacuation drill include multiple videos, each containing $\sim$100 participants who were instructed to leave through an exit as quickly as they could. This dataset are divided into three subsets based on competitiveness: Low Competitiveness (Drill$_{1}$), Moderate Competitiveness (Drill$_{2}$) and High Competitiveness (Drill$_{3}$). Hajj and Hellfest are two sets of videos from YouTube. Pilgrims in Hajj walk around the Kaaba in a counter-clockwise direction. In Hellfest, two groups of people intentionally run into one another at a high speed. The density of Evacuation, Hajj and Hellfest easily reaches 6-8$p/m^2$. Marathon is selected from UCF Dataset\cite{ali2007lagrangian}, where people run in a U-shaped road. We exclude synthetic datasets as existing simulators fail to generate realistic crowd dynamics in extremely high-density scenarios~\cite{narain2009aggregate}. Consequently, models that can learn well on synthetic data cannot guarantee to work well on real crowds.

Similar to other differentiable physics models~\cite{gong2022fine}, our model does not need large amounts of training data, because it essentially learns a parameterized PDE. In Drill$_1$, Drill$_2$ and Drill$_3$, we randomly select 4 videos for training, 2 videos for validation and 2 video for testing. For Hajj and Marathon, we divide the video frames into training, validation and testing splits using a ratio of 60\%, 20\% and 20\% respectively.For the Hellfest dataset, we use a split ratio of 40\% for training, 30\% for validation, and 30\% for testing.  Initially, we compute pixel-wise optical flow between consecutive frames using denseflow~\cite{denseflow}. Then we discretize the image space into grids and treat every optical flow vector as a velocity vector of a particle and use the P2G transfer (introduced in \ref{sec:crowd_mpm}) to estimate the velocity field.  

\begin{table*}[t]
    % \small
    \centering
    \begin{tabular}{p{1.6 cm}| p{0.8 cm} p{0.9 cm} |p{0.8 cm} p{0.9 cm}| p{0.8 cm} p{0.9 cm}|p{0.8 cm} p{0.9 cm}| p{0.8 cm} p{0.9 cm}|p{0.8 cm} p{0.9 cm}}
    \toprule
    \multirow{2}{*}{Methods} & \multicolumn{2}{c|}{Drill$_{1}$} & \multicolumn{2}{c|}{Drill$_{2}$} & \multicolumn{2}{c|}{Drill$_{3}$} & \multicolumn{2}{c|}{Hajj} & \multicolumn{2}{c|}{Hellfest} & \multicolumn{2}{c}{Marathon}\\
    \cline{2-3}\cline{4-5}\cline{6-7}\cline{8-9}\cline{10-11}\cline{12-13}
    & Err$_{vel}$ & Err$_{flow}$ & Err$_{vel}$ & Err$_{flow}$ & Err$_{vel}$ & Err$_{flow}$ & Err$_{vel}$ & Err$_{flow}$ & Err$_{vel}$ & Err$_{flow}$ & Err$_{vel}$ & Err$_{flow}$\\
    \midrule
    PredFlow & 1.0669 & 1.6944 & 10.260 & 13.944 & 2.4555 & 3.7547 & 2246.6 &3581.2 & 8.8040 & 14.040 & 122.86 & 621.70 \\
    Baseline$_I$ & 0.7555 & 1.1455 & 1.3319 & \cellcolor{yellow!40}1.8826 & \cellcolor{yellow!40}2.1150 & \cellcolor{yellow!40}3.0985 & 0.9354 &1.2511 & \cellcolor{yellow!40}3.5151 & \cellcolor{orange!40}5.2745 & 2.8778 & 11.820\\
    PredRNNv2 & 1.3239 & 1.7329 & \cellcolor{yellow!40}1.3170 & 1.9586 & \cellcolor{orange!40}1.6613 & \cellcolor{orange!40}2.8734 & 0.6805 & 1.0022 & \cellcolor{orange!40}3.3374 & \cellcolor{yellow!40}5.4267 & 2.0510 & 10.983\\
    SimVP & 2.2364 & 2.6801 & 5.6415 & 6.8968 & 2.9760 & 4.2312 & \cellcolor{red!40}0.6212 & \cellcolor{red!40}0.9160 & 4.9703 & 7.3458 & \cellcolor{orange!40}1.6636 & \cellcolor{red!40}10.633\\
    TAU & \cellcolor{orange!40}0.5330 & \cellcolor{yellow!40}0.9466 & 2.8687 & 3.6093 & 2.6546 & 3.9093 &  \cellcolor{yellow!40}0.6791 & \cellcolor{yellow!40}0.9778 & 5.3888 & 7.8822 & \cellcolor{yellow!40}1.9015 & \cellcolor{yellow!40}10.966\\
    HINN & \cellcolor{yellow!40} 0.5618& \cellcolor{red!40}0.9224 & \cellcolor{orange!40}1.1187 & \cellcolor{orange!40}1.6603 & 2.6590 & 3.6835 & 1.1600 & 1.5402 & 7.1427 & 9.5330 & 4.3488 & 14.210\\
    \midrule
    Ours(mean) & \cellcolor{red!40}0.5284 & \cellcolor{orange!40}0.9293 & \cellcolor{red!41}1.0721 & \cellcolor{red!40}1.6558 & \cellcolor{red!40}1.6461 & \cellcolor{red!40}2.7354 & \cellcolor{orange!40}0.6591 & \cellcolor{orange!40}0.9708 & \cellcolor{red!40}3.0457 & \cellcolor{red!40}4.8823 & \cellcolor{red!40}1.4927 & \cellcolor{orange!40}10.794\\
    Ours(best) & 0.5259 & 0.9263 & 1.0708 & 1.6547 & 1.6435 & 2.7326 & 0.6502 & 0.9609 & 3.0422 & 4.8790 & 1.4581 & 10.759\\
    
    \bottomrule
    \end{tabular}
    \caption{Prediction comparison. The red, orange and yellow colors represent the top three results. Our method achieves up to 18.69\% and 12.11\% improvement in Err$_{vel}$ and Err$_{flow}$ respectively on Drill$_2$.}
    \label{tab:comparison_in_predicion}
    \vspace{-0.2cm}
\end{table*}

\paragraph{Metrics} Since forecasting high-density crowds from in-the-wild videos is a new topic, there is no established evaluation protocol and metrics to our best knowledge. Therefore, we employ both quantitative and qualitative comparison. On the raw data, we employ the Mean Square Error (MSE) on the optical flow ($Err_{flow}$). The predicted velocity field is converted into optical flow via G2P transfer (\cref{sec:crowd_mpm}). Furthermore, to evaluate the field locally, we use MSE on the velocity field ($Err_{vel}$).

\subsection{Prediction}
\begin{figure*}
    \centering
    \begin{subfigure}[b]{0.04\linewidth}
        \includegraphics[width=\textwidth]{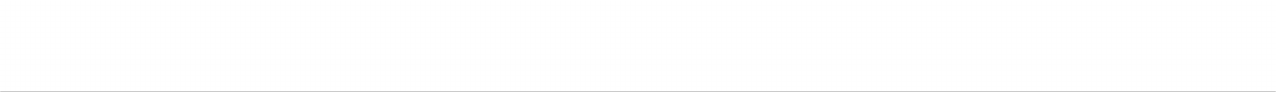}
    \end{subfigure}
    \begin{subfigure}[b]{0.58\linewidth}
        \includegraphics[width=\textwidth]{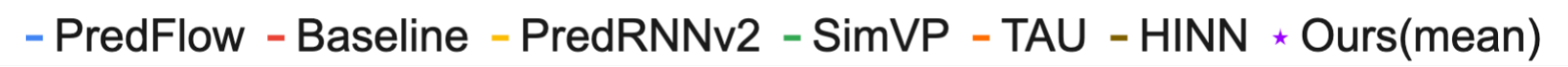}
    \end{subfigure}
    \quad
    \begin{subfigure}[b]{0.34\linewidth}
        \includegraphics[width=\textwidth]{figures/space.pdf}
    \end{subfigure}
    \begin{subfigure}[b]{1\linewidth}
        \includegraphics[width=\textwidth]{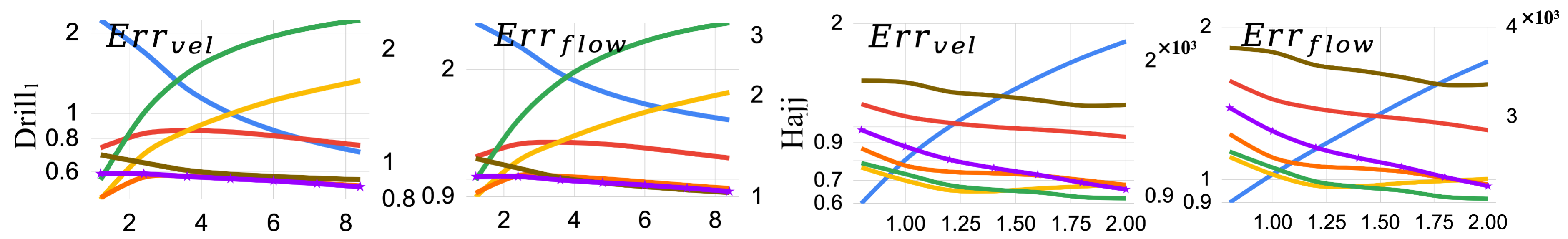}
    \end{subfigure}
    
    \begin{subfigure}[b]{1\linewidth}
        \includegraphics[width=\textwidth]{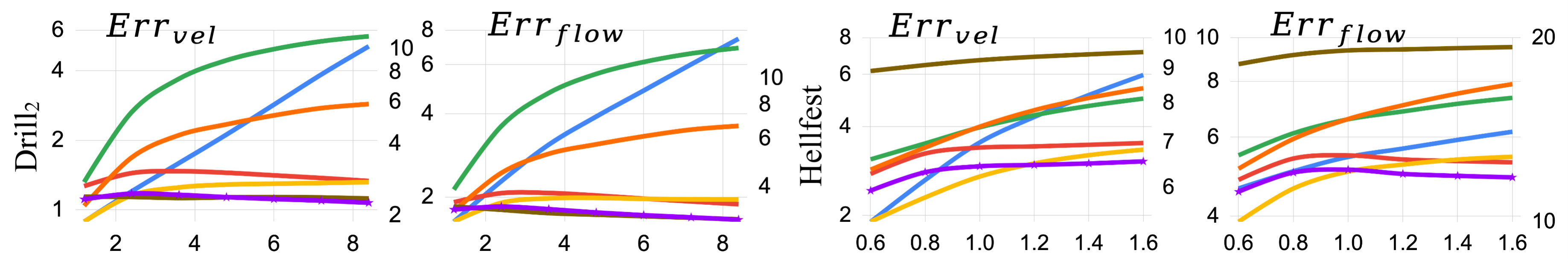}

    \end{subfigure}
    \begin{subfigure}[b]{1\linewidth}
        \includegraphics[width=\textwidth]{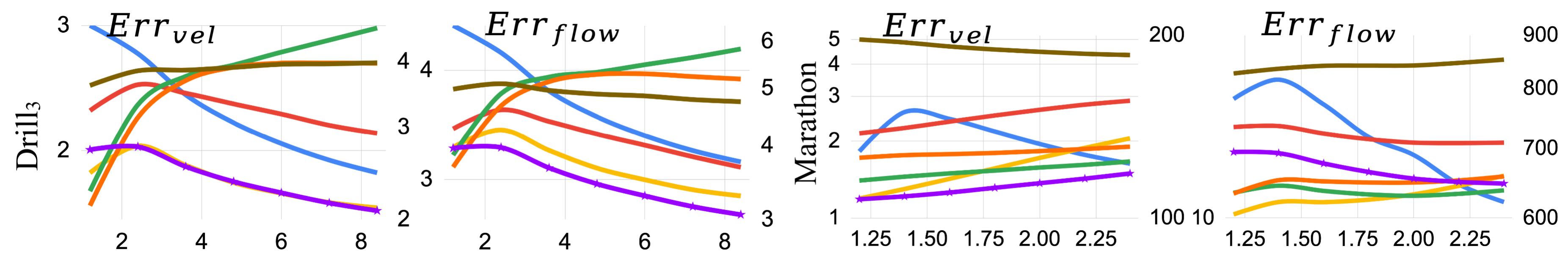}
    \end{subfigure}
    \caption{Comparison on velocity field and optical flow. The right and left vertical axes are for PredFrow and the rest methods respectively, due to the scale of PredFlow results is much larger. Vertical axes are logscaled.}
    \label{fig:predComparison}
\end{figure*}
\paragraph{Comparison baselines} 
Since there is no similar existing method, we adapt several closest methods as baselines.  They include physics models (\textbf{Baseline}$_{I}$~\cite{tampubolon2017multi}), physics-informed neural network (HINN~\cite{zhou2024hydrodynamics}), deep learning methods for scene parsing (PredFlow ~\cite{jin2017predicting}), sequence prediction (PredRNNv2~\cite{wang2022predrnn}), and video prediction (SimVP~\cite{gao2022simvp}, TAU ~\cite{tan2023temporal}). These baselines span a diverse range of architectures and tasks. We give details of how we adapt them for comparison in the SM.

Given an input up to frame $t$, we test the prediction on the frame $t+n$. If $n$ is too large, the prediction task is ill-defined as there is unobserved external impact on the crowds; if it is too small, there is too little meaningful dynamics. So after considering the total length of the data and the crowd dynamics, we choose a reasonably large $n$, corresponding to 8.2$s$ in Drill$_{1-3}$, 2$s$, 1.6$s$, and 2.4$s$ in Hajj, Hellfest and Marathon respectively.

We show the prediction results in \cref{tab:comparison_in_predicion}. Our results are the average and the best result based on 10 trials. Overall, our method achieve the highest performance across both metrics on all datasets except Hajj where our method is slightly worse than SimVP. In Drill$_1$ and Marathon, we obtain the best Err$_{vel}$ and the second best Err$_{flow}$. For further investigation, we compare predictions across varying horizons in \cref{fig:predComparison}. 
Overall, our method achieves better predictions especially in long-horizon prediction. There are two exceptions. One exception is the optical flow in Marathon. A further investigation suggests that the main reason is the main moving crowds in this scene only occupies a portion of the whole space. However, the optical flow can exist in other places due to noises. A pure data-driven approach tend to fit the optical flows in these areas better. Therefore, our method predicts the best on the velocity field but not in optical flows, as these noises are filtered in the P2G step. The other exception is Hajj. This is because Hajj contains a very peaceful crowd slowly moving around one point. Therefore, this is actually an easier prediction task, where nearly all methods perform reasonably well.

\paragraph{Continuous vs discrete time modeling} One distinctive feature of our method is that it is a continuous-time model, which does not require data to be observed on evenly discretized timeline. This is different from the other deep learning methods which are discrete-time models requiring observations to be evenly spaced by a fixed $\Delta t$. We show more results in the SM.

\paragraph{Input requirements and parameter efficiency} Our approach demonstrates a significant advantage in both input requirements and parameter efficiency. PredFlow requires four consecutive RGB frames and has 42.534M parameters, while PredRNNv2, SimVP, and TAU demand ten consecutive frames, with 24.216M, 18.604M, and 44.657M parameters, respectively. Although Baseline$_{I}$ operates on a single optical flow frame, it needs laborious hand-tuning. Our model achieves mostly the best long-horizon prediction with only a single optical flow frame as input, and merely 3.06M parameters. Vastly different from other models, our input contains little dynamics. The substantial reduction in parameter count, coupled with the minimal input requirements, highlights our model’s efficiency, especially when data and computational resources are limited.

\subsection{Prediction under Unseen Behavior}
For further generalization tests, we conduct cross-behavior testing, \ie trained on one behavior and tested on another. Specifically, we use the Drill as they are conducted in the same environment, with only the behavior changed. People were asked to be peaceful, neural and aggressive in $Drill_1$, $Drill_2$ and $Drill_3$ respectively. These scenes share some common behaviors and cross-behavior prediction can test whether a model can learn such behaviors.

\begin{figure*}
    \centering
    \begin{subfigure}[b]{0.04\linewidth}
        \includegraphics[width=\textwidth]{figures/space.pdf}
    \end{subfigure}
    \begin{subfigure}[b]{0.48\linewidth}        \includegraphics[width=\textwidth]{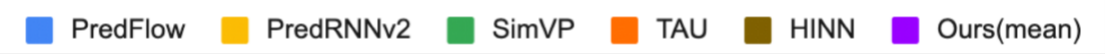}
    \end{subfigure}
    \quad
    \begin{subfigure}[b]{0.44\linewidth}
    \includegraphics[width=\textwidth]{figures/space.pdf}
    \end{subfigure}
    \begin{subfigure}[b]{1\linewidth}
        \includegraphics[width=1\linewidth]{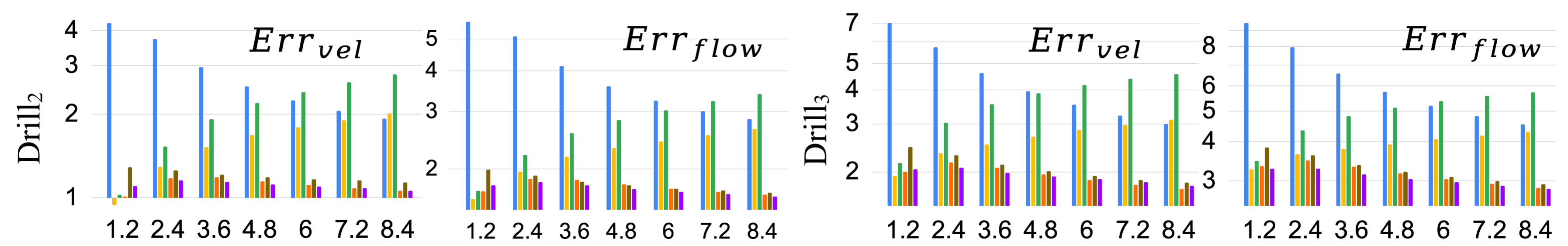}
    \end{subfigure}
    \caption{Generalization. Trained on Drill$_1$, and tested on Drill$_2$ and Drill$_3$. Vertical axes are logscaled.}
    \label{fig:genelization}
    % \vspace{-0.8mm}
\end{figure*}

Here, we conduct two groups of experiments by training the model on $Drill_1$ and test the model on $Drill_2$ and $Drill_3$. Quantitative results are reported in \cref{fig:genelization}. Obviously, as behavior shifts from peaceful to neutral and then to aggressive, model performance declines consistently across both velocity field and optical flow metrics. This suggests that the model's capacity to generalize across behaviors decreases as the level of crowd competitiveness rises, indicating that aggressive crowd dynamics present more complex interactions that are harder for the model to predict.

\subsection{Qualitative Evaluation}
One distinctive feature of our method is, once learned, it is possible to use it as a simulator for analysis, which is not possible for all the pure data-driven baseline methods. This is especially important for high-density crowd management as it often requires to predict in `what-if' situations which can be dangerous in real world and there is no data for reference. However, it is still an open question in terms of how to quantitatively evaluate the simulation of high-density crowds especially if there is no data reference. Therefore, following the practice \cite{ulicny2002towards, yang2020review}, we conduct qualitative evaluation.

\subsubsection{Crowd Material}
\begin{figure*}[tb]
\centering
\includegraphics[width=0.38\linewidth]{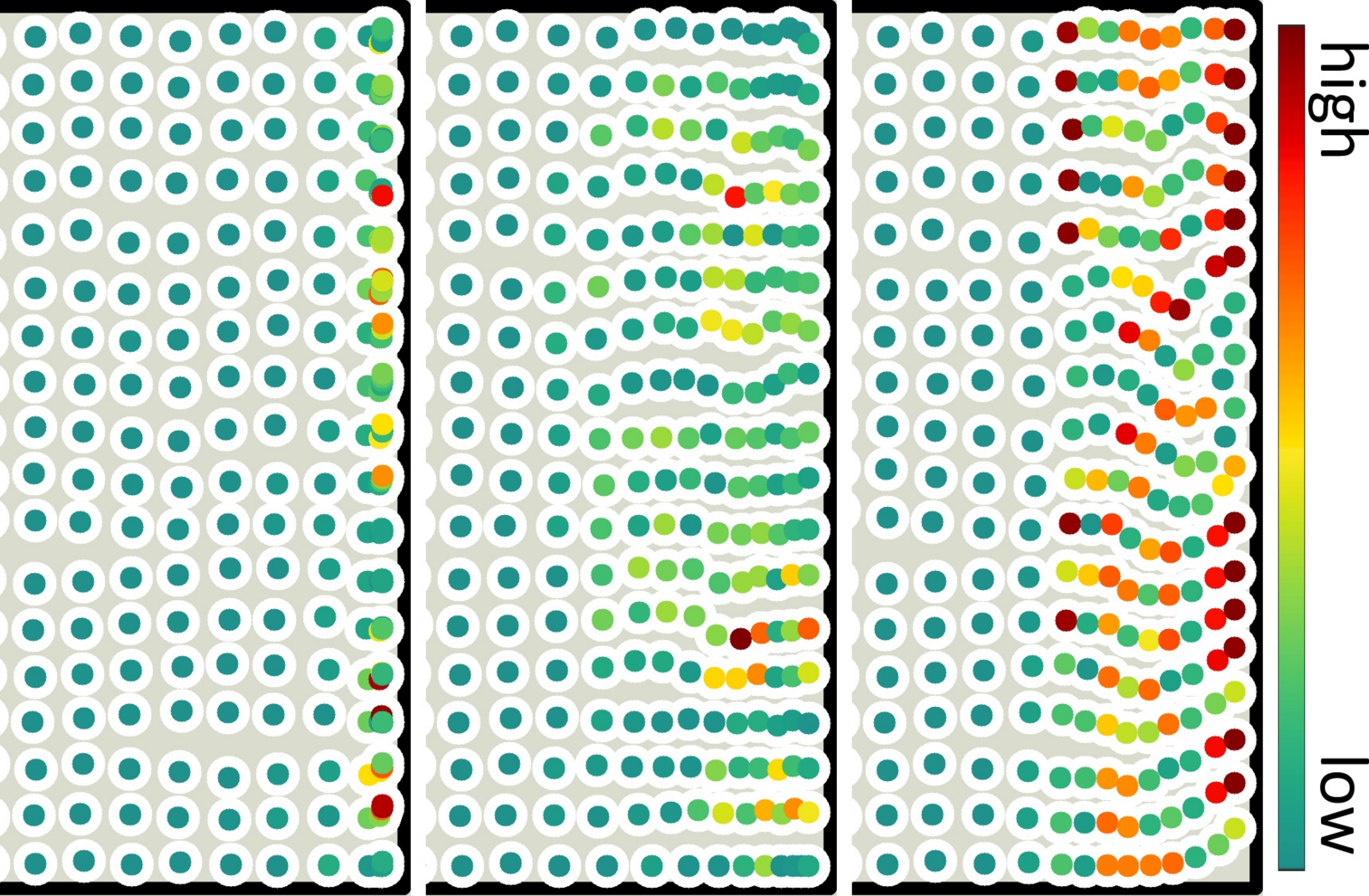}
\includegraphics[width=0.51\linewidth]{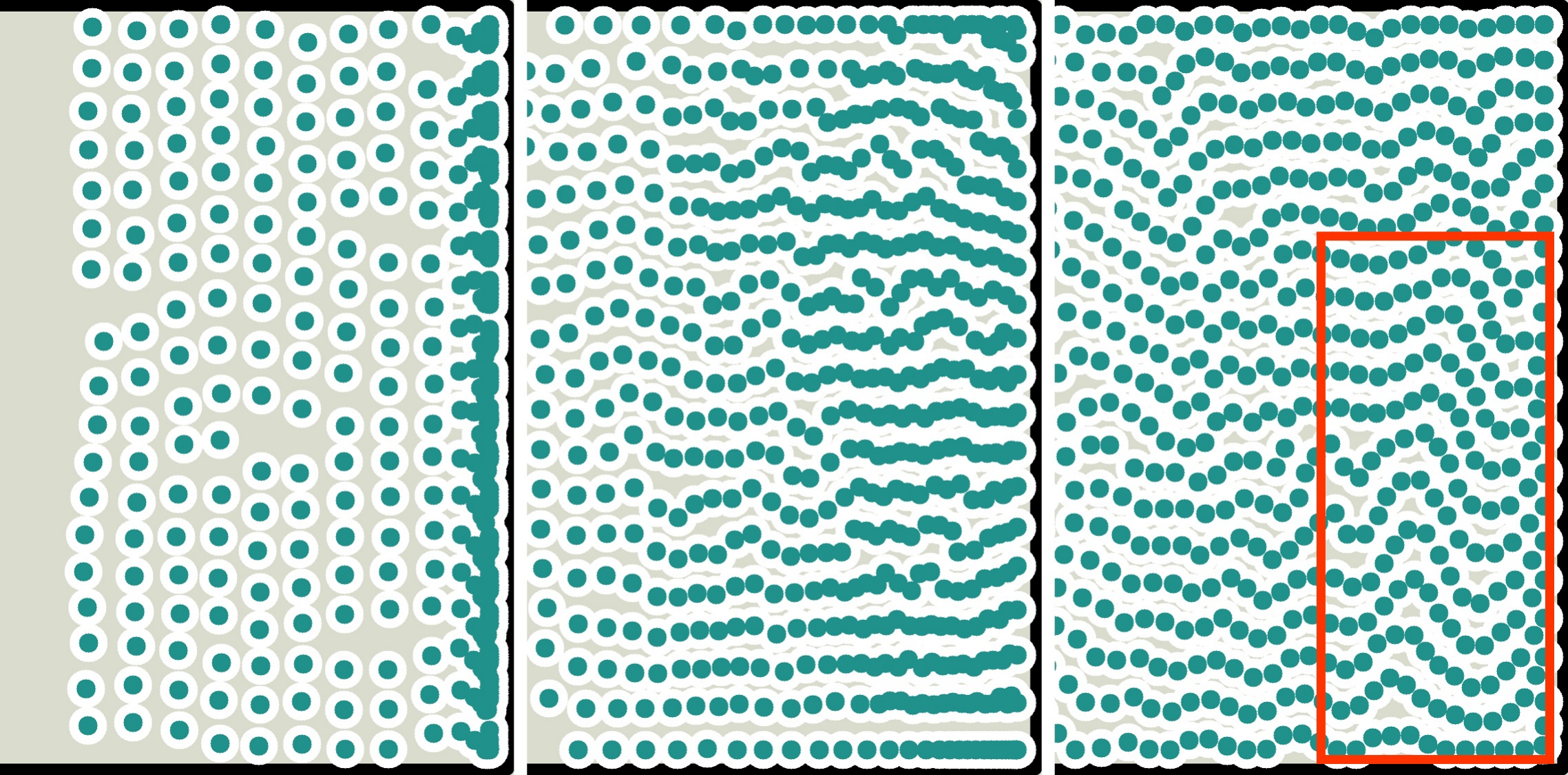}
\vspace{-0.1cm}
\caption{Crowd moving to the right with an initial velocity towards the right. Left: the stress colormap at t = 0.5s. Right: the crowd state at t = 1.8s. There are three experiments in the left and right group respectively. They are: (1) small stiffness in standard MPM $\epsilon$=1, (2) large stiffness in standard MPM $\epsilon$=100, (3) Ours.}
\label{fig:materials}
\vspace{-0.1cm}
\end{figure*}

We first design a simulation scenario where people are pushed into an enclosed area, to show how our method captures the empirical observations about `crowd material' mentioned in \cref{sec:crowd_mpm}, in comparison with a standard MPM model. It is a half-open area where people keep coming from the left and push the whole crowd to the right, shown in \cref{fig:materials}. \cref{fig:materials} Left shows the stress map. Normally, to simulate such crowds with a physics model, one needs to hand-tune the key parameters, which in standard MPM is the Young's modulus $\epsilon$. However, no matter whether the stiffness is low or high in standard MPM (\cref{fig:materials} Left 1 and 2) where the Young's modulus $\epsilon = 1$ and $\epsilon=100$, it does not model the exponential resistance as our model (\cref{fig:materials} Left 3). The overall stress increases steeply near the right end where the inter-person distance becomes smaller. This indicates even hand-tuning MPM will not mimic the crowd material as it does not exist in its parameter space. 

Next, \cref{fig:materials} Right shows the compression dominance. \cref{fig:materials} Right 1 is too soft to resist compression, while \cref{fig:materials} Right 2 resists compression, shear and rotation all together. Comparatively, our model (\cref{fig:materials} Right 3) resists compression between particles but still allows shearing/rotation as highlighted in the red box. Again, this shows that the crowd material does not exist in the parameter space of standard MPM. Furthermore, the red box shows a `wave' formation due to compression induced by the left side. Simulating wave formation is only recently addressed by ~\cite{van2021sph}, but ~\cite{van2021sph} is not learnable, unlike our method.

\subsubsection{Simulation in a New Environment}
We also show simulation where the behaviors are first learned in one environment but simulated with changes to the environment. This is a common scenario in designing a public space for public safety.

\begin{figure}
    \centering
    \includegraphics[width=0.49\textwidth]{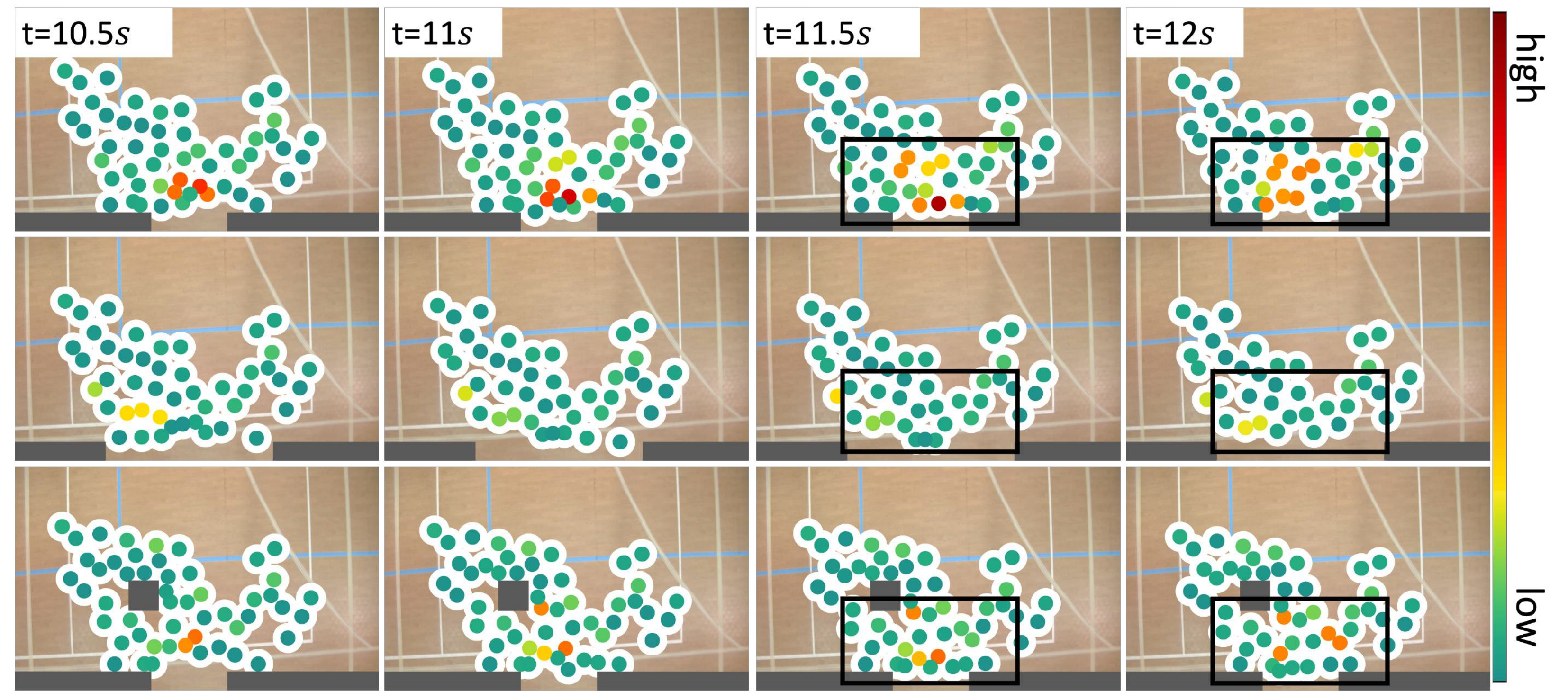}
    \caption{Evacuation simulations in the original scene (top), with a wider exit (middle) and a pillar (bottom). The color indicates the magnitude of stress.}
    \label{fig:sim_drill}
\end{figure}

We show several examples in \cref{fig:sim_drill}, where the first row shows the simulation on the original Drill$_2$ scene, the second and third rows are two
different scenarios, one with a wider exit (second row) and one with a pillar in
front of the exit (third row). Intuitively, a wider exit reduces the stress at the exit. This is shown in the second row, showing our method simulates the correct behaviors. More interestingly, an obstacle in front of the exit, if placed well, can also reduce the stress (third row). This is observed in some circumstance \eg a certain exit width and crowd density~\cite{axblad2018evacuation}. We are able to replicate similar effects. These simulations
demonstrate that the behavior learned by our method can be effectively used for simulating alternative scenarios.

\section{Discussion, Conclusion and Future Work}
\label{sec:conclusion}
In this paper, we introduced a neural stochastic differential equation system to learn extreme high-density crowds as active matter. Our model is specifically designed for high-density crowds and is not suitable for low-density crowds. In crowd modeling, it is common practice that different densities require us to choose vastly different physical models. To our knowledge, no unified model exists for all crowd densities. Our method fills the gap of learning physical models for extremely high-density crowds, which is first of its kind, to the best of our knowledge. Additionally, we focus on scenarios involving large crowd movement, as slow or static crowds are typically less critical from a safety perspective. 

While our model shows strong capabilities in capturing high-density crowd dynamics, it has certain limitations. A primary limitation is that its current single-view video input design restricts cross-view generalization, which can be improved by incorporating multiview perspectives when estimating the velocity field. Another challenge is that it does not explicitly model complex behaviors like queuing, grouping, lane formation, nor account for other space users like vehicles. These limitations can be mitigated by introducing additional inter-particle forces (\eg., grouping forces) or object detection modules. Finally, predicting entirely unseen scenes and crowds remains an open challenge. In the future, we will further generalize the model to incorporate the aforementioned scenarios, aiming for a wider range of real-world applications.

\section*{Acknowlegement}
This project has received funding from the European Union’s Horizon 2020 research and innovation program under Grant Agreement No 899739 CrowdDNA.

{
    \small
    \bibliographystyle{ieeenat_fullname}
    \bibliography{egbib}

\begin{thebibliography}{76}
\providecommand{\natexlab}[1]{#1}
\providecommand{\url}[1]{\texttt{#1}}
\expandafter\ifx\csname urlstyle\endcsname\relax
  \providecommand{\doi}[1]{doi: #1}\else
  \providecommand{\doi}{doi: \begingroup \urlstyle{rm}\Url}\fi

\bibitem[Abousamra et~al.(2021)Abousamra, Hoai, Samaras, and Chen]{abousamra2021localization}
Shahira Abousamra, Minh Hoai, Dimitris Samaras, and Chao Chen.
\newblock Localization in the crowd with topological constraints.
\newblock In \emph{Proceedings of the AAAI Conference on Artificial Intelligence}, pages 872--881, 2021.

\bibitem[Adrian et~al.(2020)Adrian, Seyfried, and Sieben]{adrian2020crowds}
Juliane Adrian, Armin Seyfried, and Anna Sieben.
\newblock {C}rowds in front of bottlenecks at entrances from the perspective of physics and social psychology.
\newblock \emph{Interface}, 17\penalty0 (165):\penalty0 20190871 --, 2020.

\bibitem[Alahi et~al.(2016)Alahi, Goel, Ramanathan, Robicquet, Fei-Fei, and Savarese]{alahi2016social}
Alexandre Alahi, Kratarth Goel, Vignesh Ramanathan, Alexandre Robicquet, Li Fei-Fei, and Silvio Savarese.
\newblock Social lstm: Human trajectory prediction in crowded spaces.
\newblock In \emph{Proceedings of the IEEE conference on computer vision and pattern recognition}, pages 961--971, 2016.

\bibitem[Ali(2008)]{ali2008taming}
Saad Ali.
\newblock \emph{Taming crowded visual scenes}.
\newblock University of Central Florida, 2008.

\bibitem[Ali and Shah(2007)]{ali2007lagrangian}
Saad Ali and Mubarak Shah.
\newblock A lagrangian particle dynamics approach for crowd flow segmentation and stability analysis.
\newblock In \emph{2007 IEEE Conference on Computer Vision and Pattern Recognition}, pages 1--6. IEEE, 2007.

\bibitem[Ali and Shah(2008)]{ali2008floor}
Saad Ali and Mubarak Shah.
\newblock Floor fields for tracking in high density crowd scenes.
\newblock In \emph{European conference on computer vision}, pages 1--14. Springer, 2008.

\bibitem[Axblad and Ortega~Gonzalez(2018)]{axblad2018evacuation}
Tom Axblad and Alvaro Ortega~Gonzalez.
\newblock Evacuation with obstacles in real-time using crowd simulation, 2018.

\bibitem[Cai et~al.(2006)Cai, Freitas, and Little]{cai2006robust}
Yizheng Cai, Nando~de Freitas, and James~J Little.
\newblock Robust visual tracking for multiple targets.
\newblock In \emph{European conference on computer vision}, pages 107--118. Springer, 2006.

\bibitem[Chen et~al.(2024)Chen, Ding, Li, Wang, and Zhang]{chen2024social}
Hongyi Chen, Jingtao Ding, Yong Li, Yue Wang, and Xiao-Ping Zhang.
\newblock Social physics informed diffusion model for crowd simulation.
\newblock In \emph{Proceedings of the AAAI Conference on Artificial Intelligence}, pages 474--482, 2024.

\bibitem[Day(1990)]{day1990no}
Michael~A Day.
\newblock The no-slip condition of fluid dynamics.
\newblock \emph{Erkenntnis}, 33\penalty0 (3):\penalty0 285--296, 1990.

\bibitem[Dreamerr86(2013)]{hajj}
Dreamerr86.
\newblock Azan \& muslim pilgrims "hajj" prayer in mecca kaaba.
\newblock \url{https://www.youtube.com/watch?v=KGukAoiGhZU&t=5s}, 2013.

\bibitem[Gao et~al.(2022)Gao, Tan, Wu, and Li]{gao2022simvp}
Zhangyang Gao, Cheng Tan, Lirong Wu, and Stan~Z Li.
\newblock Simvp: Simpler yet better video prediction.
\newblock In \emph{Proceedings of the IEEE/CVF conference on computer vision and pattern recognition}, pages 3170--3180, 2022.

\bibitem[Garcimart{\'\i}n et~al.(2016)Garcimart{\'\i}n, Parisi, Pastor, Mart{\'\i}n-G{\'o}mez, and Zuriguel]{garcimartin2016flow}
Angel Garcimart{\'\i}n, Daniel~R Parisi, Jose~M Pastor, C{\'e}sar Mart{\'\i}n-G{\'o}mez, and Iker Zuriguel.
\newblock Flow of pedestrians through narrow doors with different competitiveness.
\newblock \emph{Journal of Statistical Mechanics: Theory and Experiment}, 2016\penalty0 (4):\penalty0 043402, 2016.

\bibitem[Golas et~al.(2014)Golas, Narain, and Lin]{golas2014continuum}
Abhinav Golas, Rahul Narain, and Ming~C Lin.
\newblock Continuum modeling of crowd turbulence.
\newblock \emph{Physical review E}, 90\penalty0 (4):\penalty0 042816, 2014.

\bibitem[Gong et~al.(2022)Gong, Zhu, Bulpitt, and Wang]{gong2022fine}
Deshan Gong, Zhanxing Zhu, Andrew~J Bulpitt, and He Wang.
\newblock Fine-grained differentiable physics: a yarn-level model for fabrics.
\newblock \emph{arXiv preprint arXiv:2202.00504}, 2022.

\bibitem[Gupta et~al.(2018)Gupta, Johnson, Fei-Fei, Savarese, and Alahi]{gupta2018social}
Agrim Gupta, Justin Johnson, Li Fei-Fei, Silvio Savarese, and Alexandre Alahi.
\newblock Social gan: Socially acceptable trajectories with generative adversarial networks.
\newblock In \emph{Proceedings of the IEEE conference on computer vision and pattern recognition}, pages 2255--2264, 2018.

\bibitem[He et~al.(2020{\natexlab{a}})He, Xia, Zhao, and Wang]{He_Informative_2020}
Feixiang He, Yuanhang Xia, Xio Zhao, and He Wang.
\newblock Informative scene decomposition for crowd analysis, comparison and simulation guidance.
\newblock \emph{ACM Transaction on Graphics (TOG)}, 4\penalty0 (39), 2020{\natexlab{a}}.

\bibitem[He et~al.(2020{\natexlab{b}})He, Xiang, Zhao, and Wang]{he2020informative}
Feixiang He, Yuanhang Xiang, Xi Zhao, and He Wang.
\newblock Informative scene decomposition for crowd analysis, comparison and simulation guidance.
\newblock \emph{ACM Transactions on Graphics (TOG)}, 39\penalty0 (4):\penalty0 50--1, 2020{\natexlab{b}}.

\bibitem[He et~al.(2013)He, Yang, Chen, Gu, and Pan]{he2013review}
Gao-qi He, Yu Yang, Zhi-hua Chen, Chun-hua Gu, and Zhi-geng Pan.
\newblock A review of behavior mechanisms and crowd evacuation animation in emergency exercises.
\newblock \emph{Journal of Zhejiang University SCIENCE C}, 14\penalty0 (7):\penalty0 477--485, 2013.

\bibitem[He{\"\i}geas et~al.(2010)He{\"\i}geas, Luciani, Thollot, and Castagn{\'e}]{heigeas2010physically}
Laure He{\"\i}geas, Annie Luciani, Joelle Thollot, and Nicolas Castagn{\'e}.
\newblock A physically-based particle model of emergent crowd behaviors.
\newblock \emph{arXiv preprint arXiv:1005.4405}, 2010.

\bibitem[Helbing and Molnar(1995)]{helbing1995social}
Dirk Helbing and Peter Molnar.
\newblock Social force model for pedestrian dynamics.
\newblock \emph{Physical review E}, 51\penalty0 (5):\penalty0 4282, 1995.

\bibitem[Helbing et~al.(2007)Helbing, Johansson, and Al-Abideen]{helbing2007dynamics}
Dirk Helbing, Anders Johansson, and Habib~Zein Al-Abideen.
\newblock Dynamics of crowd disasters: An empirical study.
\newblock \emph{Phys. Rev. E}, 75:\penalty0 046109, 2007.

\bibitem[HoToox(2014)]{hellfest}
HoToox.
\newblock Wall of death (extreme) - with full force 2014.
\newblock \url{https://www.youtube.com/watch?v=ySPlanMCmM4}, 2014.

\bibitem[Hu et~al.(2008)Hu, Ali, and Shah]{hu2008learning}
Min Hu, Saad Ali, and Mubarak Shah.
\newblock Learning motion patterns in crowded scenes using motion flow field.
\newblock In \emph{2008 19th International Conference on Pattern Recognition}, pages 1--5. IEEE, 2008.

\bibitem[Hu et~al.(2018)Hu, Fang, Ge, Qu, Zhu, Pradhana, and Jiang]{hu2018moving}
Yuanming Hu, Yu Fang, Ziheng Ge, Ziyin Qu, Yixin Zhu, Andre Pradhana, and Chenfanfu Jiang.
\newblock A moving least squares material point method with displacement discontinuity and two-way rigid body coupling.
\newblock \emph{ACM Transactions on Graphics (TOG)}, 37\penalty0 (4):\penalty0 1--14, 2018.

\bibitem[Jiang et~al.(2015)Jiang, Schroeder, Selle, Teran, and Stomakhin]{jiang2015affine}
Chenfanfu Jiang, Craig Schroeder, Andrew Selle, Joseph Teran, and Alexey Stomakhin.
\newblock The affine particle-in-cell method.
\newblock \emph{ACM Transactions on Graphics (TOG)}, 34\penalty0 (4):\penalty0 1--10, 2015.

\bibitem[Jiang et~al.(2016)Jiang, Schroeder, Teran, Stomakhin, and Selle]{jiang2016material}
Chenfanfu Jiang, Craig Schroeder, Joseph Teran, Alexey Stomakhin, and Andrew Selle.
\newblock The material point method for simulating continuum materials.
\newblock In \emph{Acm siggraph 2016 courses}, pages 1--52. 2016.

\bibitem[Jiang et~al.(2010)Jiang, Xu, Mao, Li, Xia, and Wang]{jiang2010continuum}
Hao Jiang, Wenbin Xu, Tianlu Mao, Chunpeng Li, Shihong Xia, and Zhaoqi Wang.
\newblock Continuum crowd simulation in complex environments.
\newblock \emph{Computers \& Graphics}, 34\penalty0 (5):\penalty0 537--544, 2010.

\bibitem[Jin et~al.(2017)Jin, Xiao, Shen, Yang, Lin, Chen, Jie, Feng, and Yan]{jin2017predicting}
Xiaojie Jin, Huaxin Xiao, Xiaohui Shen, Jimei Yang, Zhe Lin, Yunpeng Chen, Zequn Jie, Jiashi Feng, and Shuicheng Yan.
\newblock Predicting scene parsing and motion dynamics in the future.
\newblock \emph{Advances in neural information processing systems}, 30, 2017.

\bibitem[Junior et~al.(2010)Junior, Musse, and Jung]{junior2010crowd}
Julio Cezar Silveira~Jacques Junior, Soraia~Raupp Musse, and Claudio~Rosito Jung.
\newblock Crowd analysis using computer vision techniques.
\newblock \emph{IEEE Signal Processing Magazine}, 27\penalty0 (5):\penalty0 66--77, 2010.

\bibitem[Karamouzas et~al.(2018)Karamouzas, Sohre, Hu, and Guy]{karamouzas2018crowd}
Ioannis Karamouzas, Nick Sohre, Ran Hu, and Stephen~J Guy.
\newblock Crowd space: a predictive crowd analysis technique.
\newblock \emph{ACM Transactions on Graphics (TOG)}, 37\penalty0 (6):\penalty0 1--14, 2018.

\bibitem[Lee et~al.(2007)Lee, Choi, Hong, and Lee]{lee2007group}
Kang~Hoon Lee, Myung~Geol Choi, Qyoun Hong, and Jehee Lee.
\newblock Group behavior from video: a data-driven approach to crowd simulation.
\newblock In \emph{Proceedings of the 2007 ACM SIGGRAPH/Eurographics symposium on Computer animation}, pages 109--118, 2007.

\bibitem[Liang et~al.(2022)Liang, Xu, and Bai]{liang2022end}
Dingkang Liang, Wei Xu, and Xiang Bai.
\newblock An end-to-end transformer model for crowd localization.
\newblock In \emph{Computer Vision--ECCV 2022: 17th European Conference, Tel Aviv, Israel, October 23--27, 2022, Proceedings, Part I}, pages 38--54. Springer, 2022.

\bibitem[Liang et~al.(2019)Liang, Lin, and Koltun]{liang2019differentiable}
Junbang Liang, Ming Lin, and Vladlen Koltun.
\newblock Differentiable cloth simulation for inverse problems.
\newblock \emph{Advances in Neural Information Processing Systems}, 32, 2019.

\bibitem[Liu et~al.(2023)Liu, Lu, Cao, and Liu]{liu2023point}
Chengxin Liu, Hao Lu, Zhiguo Cao, and Tongliang Liu.
\newblock Point-query quadtree for crowd counting, localization, and more.
\newblock In \emph{Proceedings of the IEEE/CVF International Conference on Computer Vision}, pages 1676--1685, 2023.

\bibitem[Liu et~al.(2024)Liu, Li, Qi, Han, van~den Hengel, Sebe, Yang, and Huang]{liu2024consistency}
Xinyan Liu, Guorong Li, Yuankai Qi, Zhenjun Han, Anton van~den Hengel, Nicu Sebe, Ming-Hsuan Yang, and Qingming Huang.
\newblock Consistency-aware anchor pyramid network for crowd localization.
\newblock \emph{IEEE transactions on pattern analysis and machine intelligence}, 2024.

\bibitem[L{\'o}pez et~al.(2019)L{\'o}pez, Chaumette, Marchand, and Pettr{\'e}]{lopez2019character}
Axel L{\'o}pez, Fran{\c{c}}ois Chaumette, Eric Marchand, and Julien Pettr{\'e}.
\newblock Character navigation in dynamic environments based on optical flow.
\newblock In \emph{Computer Graphics Forum}, pages 181--192. Wiley Online Library, 2019.

\bibitem[Marana et~al.(1998)Marana, Velastin, Costa, and Lotufo]{marana1998automatic}
A~NSAV Marana, Sergio~A Velastin, L~da~F Costa, and RA Lotufo.
\newblock Automatic estimation of crowd density using texture.
\newblock \emph{Safety Science}, 28\penalty0 (3):\penalty0 165--175, 1998.

\bibitem[Mo et~al.(2024)Mo, Fu, and Di]{mo2024pi}
Zhaobin Mo, Yongjie Fu, and Xuan Di.
\newblock Pi-neugode: Physics-informed graph neural ordinary differential equations for spatiotemporal trajectory prediction.
\newblock In \emph{Proceedings of the 23rd International Conference on Autonomous Agents and Multiagent Systems}, pages 1418--1426, 2024.

\bibitem[Narain et~al.(2009)Narain, Golas, Curtis, and Lin]{narain2009aggregate}
Rahul Narain, Abhinav Golas, Sean Curtis, and Ming~C Lin.
\newblock Aggregate dynamics for dense crowd simulation.
\newblock In \emph{ACM SIGGRAPH Asia 2009 papers}, pages 1--8. 2009.

\bibitem[Pin{\c{c}}e et~al.(2016)Pin{\c{c}}e, Velu, Callegari, Elahi, Gigan, Volpe, and Volpe]{pincce2016disorder}
Er{\c{c}}a{\u{g}} Pin{\c{c}}e, Sabareesh~KP Velu, Agnese Callegari, Parviz Elahi, Sylvain Gigan, Giovanni Volpe, and Giorgio Volpe.
\newblock Disorder-mediated crowd control in an active matter system.
\newblock \emph{Nature communications}, 7\penalty0 (1):\penalty0 10907, 2016.

\bibitem[Raissi et~al.(2019)Raissi, Perdikaris, and Karniadakis]{raissi2019physics}
Maziar Raissi, Paris Perdikaris, and George~E Karniadakis.
\newblock Physics-informed neural networks: A deep learning framework for solving forward and inverse problems involving nonlinear partial differential equations.
\newblock \emph{Journal of Computational physics}, 378:\penalty0 686--707, 2019.

\bibitem[Shen et~al.(2021)Shen, Yin, Shao, Wang, Jiang, Lan, and Zhou]{shen2021high}
Siyuan Shen, Yang Yin, Tianjia Shao, He Wang, Chenfanfu Jiang, Lei Lan, and Kun Zhou.
\newblock High-order differentiable autoencoder for nonlinear model reduction.
\newblock \emph{arXiv preprint arXiv:2102.11026}, 2021.

\bibitem[Sieben and Seyfried(2023)]{sieben2023inside}
Anna Sieben and Armin Seyfried.
\newblock Inside a life-threatening crowd: Analysis of the love parade disaster from the perspective of eyewitnesses.
\newblock \emph{Safety Science}, 166:\penalty0 106229, 2023.

\bibitem[Singh et~al.(2020)Singh, Rajora, Vishwakarma, Tripathi, Kumar, and Walia]{singh2020crowd}
Kuldeep Singh, Shantanu Rajora, Dinesh~Kumar Vishwakarma, Gaurav Tripathi, Sandeep Kumar, and Gurjit~Singh Walia.
\newblock Crowd anomaly detection using aggregation of ensembles of fine-tuned convnets.
\newblock \emph{Neurocomputing}, 371:\penalty0 188--198, 2020.

\bibitem[Song et~al.(2024)Song, Wang, Yang, Taccari, and Chen]{song2024loss}
Yanjie Song, He Wang, He Yang, Maria~Luisa Taccari, and Xiaohui Chen.
\newblock Loss-attentional physics-informed neural networks.
\newblock \emph{Journal of Computational Physics}, page 112781, 2024.

\bibitem[Still()]{crowd_risky}
Prof. Dr. G.~Keith Still.
\newblock Crowd safety and crowd risk analysis.

\bibitem[Tampubolon et~al.(2017)Tampubolon, Gast, Kl{\'a}r, Fu, Teran, Jiang, and Museth]{tampubolon2017multi}
Andre~Pradhana Tampubolon, Theodore Gast, Gergely Kl{\'a}r, Chuyuan Fu, Joseph Teran, Chenfanfu Jiang, and Ken Museth.
\newblock Multi-species simulation of porous sand and water mixtures.
\newblock \emph{ACM Transactions on Graphics (TOG)}, 36\penalty0 (4):\penalty0 1--11, 2017.

\bibitem[Tan et~al.(2023)Tan, Gao, Wu, Xu, Xia, Li, and Li]{tan2023temporal}
Cheng Tan, Zhangyang Gao, Lirong Wu, Yongjie Xu, Jun Xia, Siyuan Li, and Stan~Z Li.
\newblock Temporal attention unit: Towards efficient spatiotemporal predictive learning.
\newblock In \emph{Proceedings of the IEEE/CVF Conference on Computer Vision and Pattern Recognition}, pages 18770--18782, 2023.

\bibitem[Temam(2001)]{temam2001navier}
Roger Temam.
\newblock \emph{Navier-Stokes equations: theory and numerical analysis}.
\newblock American Mathematical Soc., 2001.

\bibitem[Toner and Tu(1995)]{toner1995long}
John Toner and Yuhai Tu.
\newblock Long-range order in a two-dimensional dynamical xy model: how birds fly together.
\newblock \emph{Physical review letters}, 75\penalty0 (23):\penalty0 4326, 1995.

\bibitem[Treuille et~al.(2006)Treuille, Cooper, and Popovi{\'c}]{treuille2006continuum}
Adrien Treuille, Seth Cooper, and Zoran Popovi{\'c}.
\newblock Continuum crowds.
\newblock \emph{ACM Transactions on Graphics (TOG)}, 25\penalty0 (3):\penalty0 1160--1168, 2006.

\bibitem[Ulicny and Thalmann(2002)]{ulicny2002towards}
Branislav Ulicny and Daniel Thalmann.
\newblock Towards interactive real-time crowd behavior simulation.
\newblock In \emph{Computer Graphics Forum}, pages 767--775. Wiley Online Library, 2002.

\bibitem[Ummenhofer et~al.(2019)Ummenhofer, Prantl, Thuerey, and Koltun]{ummenhofer2019lagrangian}
Benjamin Ummenhofer, Lukas Prantl, Nils Thuerey, and Vladlen Koltun.
\newblock Lagrangian fluid simulation with continuous convolutions.
\newblock In \emph{International Conference on Learning Representations}, 2019.

\bibitem[Van~den Berg et~al.(2008)Van~den Berg, Lin, and Manocha]{van2008reciprocal}
Jur Van~den Berg, Ming Lin, and Dinesh Manocha.
\newblock Reciprocal velocity obstacles for real-time multi-agent navigation.
\newblock In \emph{2008 IEEE international conference on robotics and automation}, pages 1928--1935. Ieee, 2008.

\bibitem[Van~Toll and Pettr{\'e}(2021)]{van2021algorithms}
Wouter Van~Toll and Julien Pettr{\'e}.
\newblock Algorithms for microscopic crowd simulation: Advancements in the 2010s.
\newblock In \emph{Computer Graphics Forum}, pages 731--754. Wiley Online Library, 2021.

\bibitem[van Toll et~al.(2021)van Toll, Chatagnon, Braga, Solenthaler, and Pettr{\'e}]{van2021sph}
Wouter van Toll, Thomas Chatagnon, C{\'e}dric Braga, Barbara Solenthaler, and Julien Pettr{\'e}.
\newblock Sph crowds: Agent-based crowd simulation up to extreme densities using fluid dynamics.
\newblock \emph{Computers \& Graphics}, 98:\penalty0 306--321, 2021.

\bibitem[Wang and O’Sullivan(2016)]{wang2016globally}
He Wang and Carol O’Sullivan.
\newblock Globally continuous and non-markovian crowd activity analysis from videos.
\newblock In \emph{European conference on computer vision}, pages 527--544. Springer, 2016.

\bibitem[Wang et~al.(2016{\natexlab{a}})Wang, Ond{\v{r}}ej, and O'Sullivan]{wang_path_2016}
He Wang, Jan Ond{\v{r}}ej, and Carol O'Sullivan.
\newblock Path patterns: Analyzing and comparing real and simulated crowds.
\newblock In \emph{ACM SIGGRAPH Symposium on Interactive 3D Graphics and Games 2016}, pages 49--57, 2016{\natexlab{a}}.

\bibitem[Wang et~al.(2016{\natexlab{b}})Wang, Ond{\v{r}}ej, and O'Sullivan]{wang_trending_2016}
He Wang, Jan Ond{\v{r}}ej, and Carol O'Sullivan.
\newblock Trending paths: A new semantic-level metric for comparing simulated and real crowd data.
\newblock \emph{IEEE Transactions on Visualization and Computer Graphics}, PP\penalty0 (99):\penalty0 1--1, 2016{\natexlab{b}}.

\bibitem[Wang et~al.(2020)Wang, Li, Zhao, Xiong, Wang, and Lin]{denseflow}
Shiguang* Wang, Zhizhong* Li, Yue Zhao, Yuanjun Xiong, Limin Wang, and Dahua Lin.
\newblock {denseflow}.
\newblock \url{https://github.com/open-mmlab/denseflow}, 2020.

\bibitem[Wang et~al.(2018)Wang, Gao, Long, Wang, and Philip]{wang2018predrnn++}
Yunbo Wang, Zhifeng Gao, Mingsheng Long, Jianmin Wang, and S~Yu Philip.
\newblock Predrnn++: Towards a resolution of the deep-in-time dilemma in spatiotemporal predictive learning.
\newblock In \emph{International conference on machine learning}, pages 5123--5132. PMLR, 2018.

\bibitem[Wang et~al.(2022)Wang, Wu, Zhang, Gao, Wang, Philip, and Long]{wang2022predrnn}
Yunbo Wang, Haixu Wu, Jianjin Zhang, Zhifeng Gao, Jianmin Wang, S~Yu Philip, and Mingsheng Long.
\newblock Predrnn: A recurrent neural network for spatiotemporal predictive learning.
\newblock \emph{IEEE Transactions on Pattern Analysis and Machine Intelligence}, 45\penalty0 (2):\penalty0 2208--2225, 2022.

\bibitem[Werling et~al.(2021)Werling, Omens, Lee, Exarchos, and Liu]{werling2021fast}
Keenon Werling, Dalton Omens, Jeongseok Lee, Ioannis Exarchos, and C~Karen Liu.
\newblock Fast and feature-complete differentiable physics for articulated rigid bodies with contact.
\newblock \emph{arXiv preprint arXiv:2103.16021}, 2021.

\bibitem[Wolinski et~al.(2014)Wolinski, J.~Guy, Olivier, Lin, Manocha, and Pettr{\'e}]{wolinski2014parameter}
David Wolinski, S J.~Guy, A-H Olivier, Ming Lin, Dinesh Manocha, and Julien Pettr{\'e}.
\newblock Parameter estimation and comparative evaluation of crowd simulations.
\newblock In \emph{Computer Graphics Forum}, pages 303--312. Wiley Online Library, 2014.

\bibitem[Wu et~al.(2017)Wu, Yang, Zheng, Su, Fan, and Yang]{wu2017crowd}
Shuang Wu, Hua Yang, Shibao Zheng, Hang Su, Yawen Fan, and Ming-Hsuan Yang.
\newblock Crowd behavior analysis via curl and divergence of motion trajectories.
\newblock \emph{International Journal of Computer Vision}, 123\penalty0 (3):\penalty0 499--519, 2017.

\bibitem[Xia et~al.(2022)Xia, Wong, Peng, Yuan, and You]{xia2022cscnet}
Beihao Xia, Conghao Wong, Qinmu Peng, Wei Yuan, and Xinge You.
\newblock Cscnet: Contextual semantic consistency network for trajectory prediction in crowded spaces.
\newblock \emph{Pattern Recognition}, 126:\penalty0 108552, 2022.

\bibitem[Xiang et~al.(2024)Xiang, Haoteng, Wang, and Jin]{xiang2024socialcvae}
Wei Xiang, YIN Haoteng, He Wang, and Xiaogang Jin.
\newblock Socialcvae: Predicting pedestrian trajectory via interaction conditioned latents.
\newblock In \emph{Proceedings of the AAAI Conference on Artificial Intelligence}, pages 6216--6224, 2024.

\bibitem[Yang et~al.(2020)Yang, Li, Gong, Peng, and Hu]{yang2020review}
Shanwen Yang, Tianrui Li, Xun Gong, Bo Peng, and Jie Hu.
\newblock A review on crowd simulation and modeling.
\newblock \emph{Graphical Models}, 111:\penalty0 101081, 2020.

\bibitem[Yang et~al.(2009)Yang, Liu, and Shah]{yang2009video}
Yang Yang, Jingen Liu, and Mubarak Shah.
\newblock Video scene understanding using multi-scale analysis.
\newblock In \emph{2009 IEEE 12th International Conference on Computer Vision}, pages 1669--1676. IEEE, 2009.

\bibitem[{Yue} et~al.(2022){Yue}, {Manocha}, and {Wang}]{Jiang_trajectory_2022}
J. {Yue}, D. {Manocha}, and H. {Wang}.
\newblock Human trajectory prediction via neural social physics.
\newblock In \emph{Proceedings of the European Conference on Computer Vision (ECCV)}, 2022.

\bibitem[Yue et~al.(2023)Yue, Manocha, and Wang]{yue2023human}
Jiangbei Yue, Dinesh Manocha, and He Wang.
\newblock Human trajectory forecasting with explainable behavioral uncertainty, 2023.

\bibitem[Zhang et~al.(2020)Zhang, Wang, Jimack, and Wang]{Zhang_MeshingNet_2020}
Zheyan Zhang, Yongxing Wang, Peter~K. Jimack, and He Wang.
\newblock Meshingnet: A new mesh generation method based on deep learning.
\newblock In \emph{Computational Science -- ICCS 2020}, pages 186--198, Cham, 2020. Springer International Publishing.

\bibitem[Zhang et~al.(2021)Zhang, Jimack, and Wang]{zhang_meshingnet3d_2021}
Zheyan Zhang, Peter~K. Jimack, and He Wang.
\newblock {MeshingNet3D}: {Efficient} generation of adapted tetrahedral meshes for computational mechanics.
\newblock \emph{Advances in Engineering Software}, 157-158, 2021.

\bibitem[Zhong et~al.(2023)Zhong, Liang, Zharkov, and Neumann]{zhong2023mmvp}
Yiqi Zhong, Luming Liang, Ilya Zharkov, and Ulrich Neumann.
\newblock Mmvp: Motion-matrix-based video prediction.
\newblock In \emph{Proceedings of the IEEE/CVF International Conference on Computer Vision}, pages 4273--4283, 2023.

\bibitem[Zhou et~al.(2024)Zhou, Lai, Yu, Xiong, and Yang]{zhou2024hydrodynamics}
Yanshan Zhou, Pingrui Lai, Jiaqi Yu, Yingjie Xiong, and Hua Yang.
\newblock Hydrodynamics-informed neural network for simulating dense crowd motion patterns.
\newblock In \emph{Proceedings of the 32nd ACM International Conference on Multimedia}, pages 4553--4561, 2024.

\end{thebibliography}
}

\clearpage
\twocolumn[
\begin{@twocolumnfalse}
    \centering 
    \Large\textbf{Learning Extremely High Density Crowds as Active Matters \\ \vspace{0.1cm} Supplemental Material} \\ \vspace{0.25cm}
    \normalsize 
    \parbox{\textwidth}{
    \centering
    Feixiang He$^2$, Jiangbei Yue$^3$, Jialin Zhu$^2$, Armin Seyfried$^4$, Dan Casas$^5$, Julien Pettré$^6$, He Wang $^{1,2}$\\$^1$AI Centre, University College London, UK\ \ \ \ $^2$University College London, UK\\ $^3$University of Leeds, UK\ \ \ \ $^4$ Forschungszentrum J\"{u}lich, Germany \\$^5$Universidad Rey Juan Carlos, Spain\ \ \ $^6$ INRIA Rennes, France
    } \vspace{0.2cm}
\end{@twocolumnfalse}
]

\setcounter{section}{0}

\section{Additional Technical Details}
\subsection{Loss function}
During training, We employ the Mean Square Error loss between the predicted velocity field $\{\hat{\vect{v}}^t\}_{t=1}^T$ and the ground truth $\{\vect{v}^t\}_{t=0}^T$  to optimize the model:
\begin{equation}
    \text{min } \frac{1}{T}\int_{t=1}^T ||\vect{v}^t - \hat{\vect{v}}^t||_2^2 \approx \frac{1}{T}\sum_{t=1}^T ||\vect{v}^t - \hat{\vect{v}}^t||_2^2
\end{equation}
where we can only approximate the first integral as the summation over observed frames. But the nature of the approximation does not dictate how the frames are distributed on the timeline. They can be evenly or unevenly distributed.

\subsection{Error Metrics}
To evaluate our proposed model, we employ two metrics: $Err_{vel}$ and $Err_{flow}$. As presented in Sec. 4 of the main paper, $Err_{vel}$ is used to assess the velocity field locally and is calculated as:
\begin{equation}
    Err_{vel} = \frac{1}{T}\sum_{t=1}^T ||\vect{v}^t - \hat{\vect{v}}^t||_2^2
\end{equation}
Differently, $Err_{flow}$ operates directly on the raw optical flow data and is defined as:
\begin{equation}
    Err_{flow} = \frac{1}{T}\sum_{t=1}^T ||\vect{o}^t - \hat{\vect{o}}^t||_2^2
\end{equation}
Here, $\{\hat{\vect{o}}^t\}_{t=1}^T$ and $\{\vect{o}^t\}_{t=0}^T$ represent the predicted optical flow and the ground truth respectively. Note that our model predicts the velocity field instead of directly predicting optical flow. Therefore, the predicted velocity field is converted into optical flow via G2P transfer, as detailed in sec.3.2 of the main paper.

\subsection{Neural Networks for Crowd Material}
\begin{figure}[tb]
    \centering
    \includegraphics[width=0.5\textwidth]{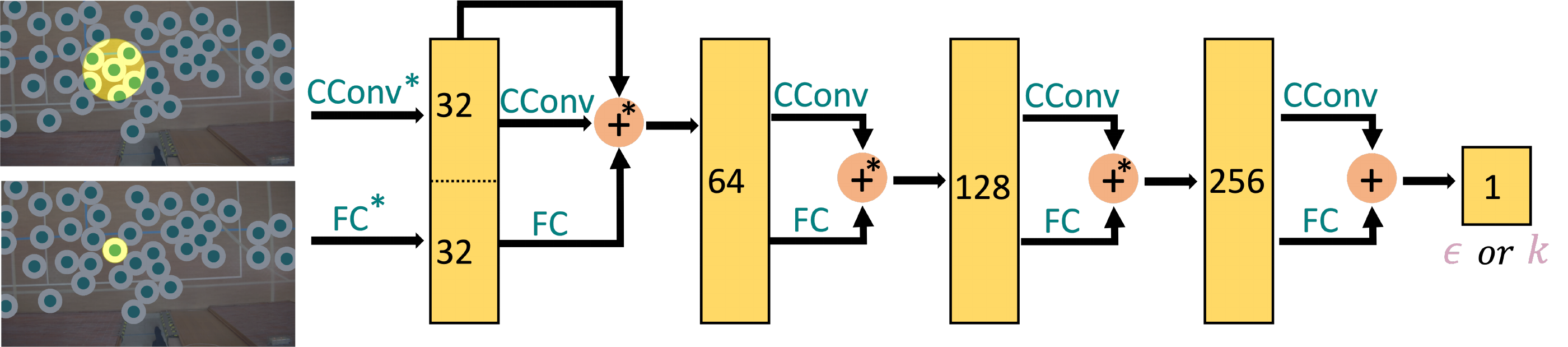}
    \caption{Parameter estimation network. Operations annotated by * are followed by a ReLU activation.}
    % All CConv and FC operations use an additive bias
    \label{fig:model}
\end{figure}

As illustrated in Sec 3.6 in the main paper,  there are two types of learnable parameters: Young's modulus $\epsilon$ and $k$. These parameters are estimated using neural networks, specifically $NN_{\epsilon}$ for $\epsilon$  and $NN_{k}$ for $k$.  Both networks share the same architecture, as shown in \cref{fig:model}. The design is mainly to consider the motions of particles and its neighboring particles. Since we model a continumm where particles can be anywhere and not necessarily evenly spaced, one of the component in the network is the Continuous Convolution or CCov, inspired by the convolution layer designed originally for a Lagrangian method ~\cite{ummenhofer2019lagrangian}. The continuous convolution captures the relative motions of the particles in the neighborhood, which dictates the local deformation of the field. In addition, we assume that the material parameter is different for every particle (pedestrian) in space and time. Therefore, we also use a fully-connected component (FC) to allow the individual particle motion to play a bigger role compared with its neighbors whose motions are already considered in the continuous convolution of the neighborhood. After several repeated layers of CCov and FC components, the network finally outputs the target parameter.

\subsection{Conditional Variational Autoencoder for Random Active Forces}
As outlined in the main paper, active force learning is divided into two parts. The first part, motion alignment (represented by $\alpha\vect{v}$), is learned through a straightforward 6-layer convolutional neural network $NN_{\alpha}$
 , where the kernel size, stride, and padding are set to 3, 1, and 1, respectively. Each layer, except the final one, is followed by a Tanh activation function. The channel configuration for the layers is 32, 64, 128, 64, 32, and 1, respectively. The input of $NN_{\alpha}$ is the velocity field at the current step, and its output is the corresponding $\alpha$.

\begin{figure}[tb]
    \centering
    \includegraphics[width=0.5\textwidth]{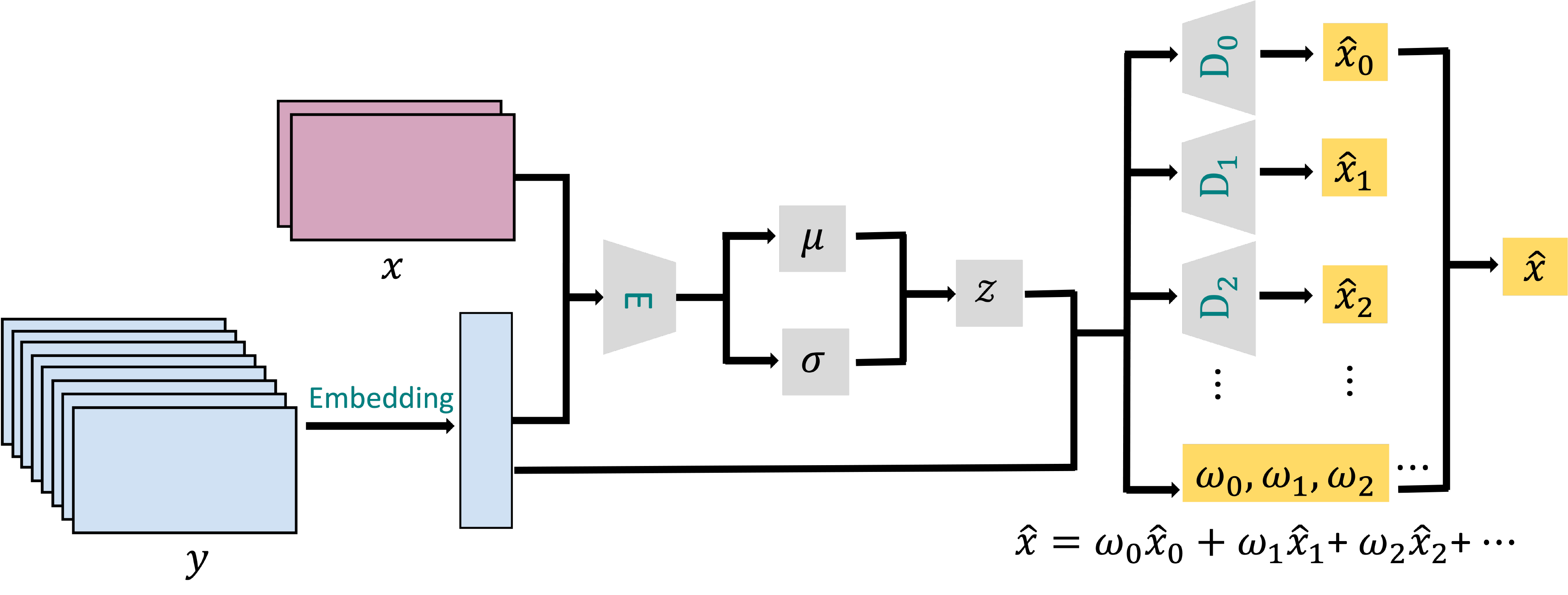}
    \caption{Architecture of CVAE. }
    \label{fig:cvae_model}
\end{figure}

Secondly, to model the rest (expressed as $- \beta|\vect{v}|^2\vect{v} + D_L\nabla(\nabla\cdot\vect{v}) +D_1\nabla^2\vect{v} + D_2(\vect{v}\cdot\nabla)^2\vect{v} + \Tilde{\vect{f}}$), a conditional variational autoencoder is utilized, as illustrated in \cref{fig:cvae_model}. The CVAE takes two inputs: $x$, representing the ground truth of the remaining active forces, and y, serving as the condition. The condition is composed of four terms $|\vect{v}|^2\vect{v}$, $\nabla(\nabla\cdot\vect{v})$, $\nabla^2\vect{v}$ and $(\vect{v}\cdot\nabla)^2\vect{v}$, and processed through an embedding layer. Note that the decoder consists of multiple components ($D_0, D_1, D_2, \cdot\cdot\cdot$) which generate intermediate outputs ($\hat{x}_0, \hat{x}_1,\hat{x}_2, \cdot\cdot\cdot$). These intermediate results are combined using a weighted sum with weights ($\omega_0, \omega_1, \omega_2, \cdot\cdot\cdot$) to produce the final output $\hat{x}$. This architecture allows the CVAE to effectively model the complex relationships between the input and the conditional data, making it suitable for learning the intricate dynamics.

\subsection{Optical flows to/from grid}
Optical flows can be transferred to and from a grid representation to for training, testing, analysis and comparison. When transferring optical flow data to a grid, we make use of the particle-to-grid operation in the MPM (P2G). Each pixel in the optical flow field is treated as an individual particle, with its velocity mapped to the corresponding grid cell using a B-spline function. Conversely, when transforming the velocity field from a grid back to its corresponding optical flows, we use the grid-to-particle operation in the MPM (G2P). G2P computes the velocity at grid nodes back to each pixel, which is treated as the optical flow at that pixel.  

This P2G then G2P mapping between optical flows and the velocity field, aside from facilitate learning, has a smoothening effect due to the B-spline based interpolation, effectively removing random noises from the noisy optical flows. In addition, this mapping also preserve the spatial and temporal dynamics. This dual representation enables the comparisons of the particle-level details and the grid-level velocity.

\subsection{Hyperparameters}
In all datasets, the density is fixed at 1. This is because it is not possible estimate each individual's mass, and if it were, there is no commonly agreed way of defining the density in high-density crowds ~\cite{marana1998automatic}. By defining the density to be 1, the particle mass depends on the particle radius. The particle radii are set to 20 pixels on $Drill_{1-3}$, 5 pixels on $Hajj$ and $Hellfest$, 10 pixels on $Marathon$. Note this setting will not affect the learning of the dynamics as the dynamics are learned mainly through the material parameters and the active force.

For training, we employ Adam optimizer with an initial learning rate of $1\times10^{-4}$. To dynamically adjust the learning rate during training, we apply a LambdaLR scheduler with a decay function defined as $ lr = 1\times 10^{-4} \times  0.9^{epoch/50}$, epoch is the current iteration. The batch size is set to 4.

% \subsection{Body forces}

\subsection{Adaptation of baseline methods}
To demonstrate the superiority of out method, we adapt several closest methods as baselines. \textbf{Baseline}$_{I}$ is a physics-based method that models crowds using a weakly incompressible fluid framework~\cite{tampubolon2017multi}. PredFlow ~\cite{jin2017predicting} is a deep learning method designed for predicting future scene parsing and motion dynamics. Additionally, we include several sequence prediction models, PredRNNv2~\cite{wang2022predrnn}, SimVP~\cite{gao2022simvp}, and TAU ~\cite{tan2023temporal},  which are also pure deep learning approaches and intended for video prediction tasks. PredFlow takes four consecutive RGB frames as input and outputs the future optical flows in an autoregressive fashion, \textbf{Baseline}$_{I}$, however, simulates velocity fields but we need to exhaustively tune the parameters. For the video prediction methods, we adapt them to take optical flow as input and forecast future optical flow sequences. Specifically, during the training, we take 1.2$s$ optical flow as input and forecast the next 1.2$s$ optical flow. These baselines span a diverse range of architectures, from physics-based models to cutting-edge deep learning frameworks, ensuring comprehensive coverage across methodologies. 

\subsection{Simulation}
Our simulation runs $\sim$ 40 FPS for 75 individuals. Like any simulator, our method experiences a decrease in performance when scaling to a large number of agents. Also, albeit now shown in the paper, our method allows dynamic adding or removing objects (dynamic or static), environmental changes or events (\eg. explosion, evacuation). The first two can be easily handled by boundary conditions, while the latter can be realized by adding particle forces to repel people from certain places, \eg. explosion, or attract them, \eg. evacuation.

\begin{table*}[tb]
    % \small
    \centering
    \begin{tabular}{p{1.0 cm}| p{0.70 cm} p{0.70 cm} p{0.70 cm} p{0.70 cm} p{0.70 cm} p{0.70 cm} p{0.8 cm} | p{0.70 cm} p{0.70 cm} p{0.70 cm} p{0.70 cm} p{0.70 cm} p{0.70 cm} p{0.70 cm}}
    \toprule
    \multirow{2}{*}{Percent} & \multicolumn{7}{c|}{Err$_{vel}$} & \multicolumn{7}{c}{Err$_{flow}$} \\
    \cline{2-8}\cline{8-15}
    & 1.2$s$ & 2.4$s$ & 3.6$s$ & 4.8$s$ & 6.0$s$ & 7.2$s$ & 8.4$s$ & 1.2$s$ & 2.4$s$ & 3.6$s$ & 4.8$s$ & 6.0$s$ & 7.2$s$ & 8.4$s$ \\
    \midrule

    0\% & 1.1099 &1.1748 & 1.1600 & 1.1332 & 1.1128 & 1.0954 & 1.0721 & 1.8016 & 1.8463 & 1.8072 & 1.7574& 1.7185 & 1.6883 & 1.6558 \\
    50\% & 1.1078 & 1.1632 & 1.1419 & 1.1155 & 1.0966 & 1.0850 & 1.0640 & 1.7958 & 1.8303 & 1.7843 & 1.7349 & 1.6974 & 1.6729 & 1.6435\\
    70\% &1.1121 & 1.1637 & 1.1381 &1.1069 &1.0853 & 1.0705 & 1.0729& 1.8086 & 1.8386& 1.7891& 1.7339 & 1.6942& 1.6653& 1.6378 \\
    
    \bottomrule
    \end{tabular}
    \caption{Evaluation on temporal discontinuities on $Drill_2$.}
    \label{tab:evaluation_on_discontinuities}
    \vspace{-0.2cm}
\end{table*}

\section{Additional Experimental Results}
\subsection{Continuous-time Prediction}
As aforementioned in the main paper, one distinctive feature of our method is that it is a continuous-time model, which does not require data to be observed on evenly discretized timeline. This is because the core of our method is a PDE which can be solved for any horizon in future. This feature provides good flexibility in terms of data collection. For CCTV camera videos, sometimes there can be objects \eg tree leaves flying across the view or certain sudden change of the lighting condition. Under these conditions, the optical flows will be extremely unreliable. This requires certain data pre-processing step, \eg abandoning frames that are of too low quality. However, in this situation, the other baseline methods will not be able to be directly trained on the pre-processed data, while our method can still be trained on data unevenly distributed on the timeline. We refer to data unevenly distributed in time as \textit{temporal discontinuities}.

To verify the robustness of our model under temporal discontinuities, we conduct three groups of experiments by randomly masking 0\%, 50\% and 70\% of the frames respectively. Accordingly, we also can easily adapt our loss function to:
\begin{equation}
    \text{min } \frac{1}{T}\int_{t=1}^T ||\vect{v}^t - \hat{\vect{v}}^t||_2^2 \approx \frac{1}{T}\sum_{t\in\Phi} ||\vect{v}^t - \hat{\vect{v}}^t||_2^2
\end{equation}
where $\Phi$ is the set of the observed frames and $T=|\Phi|$.

The corresponding experimental results are presented in \cref{tab:evaluation_on_discontinuities}. For both velocity prediction errors ($Err_{vel}$) and flow prediction errors ($Err_{flow}$), the performance remains stable across varying masking levels, with only minor variations observed. Notably, even under the 70\% masking scenario, the errors are comparable to those without masking. These findings highlight the model's capacity to maintain accuracy and adaptability despite significant temporal discontinuities, making it well-suited for real-world applications where data observations are sparse or unevenly distributed in time.

\subsection{Robustness to Input Noise}

\begin{table}[t]
    \vspace{-3mm}
    \centering
    \footnotesize
    \begin{minipage}[t]{0.14\textwidth}
    \begin{tabular}{p{0.2cm}|p{0.5cm} p{0.6cm}}
    \toprule
    Std & Err$_{vel}$ & Err$_{flow}$ \\
    \midrule
    0.1 & 1.0437 & 1.6281 \\
    1 & 1.0615& 1.6438 \\
    10 & 1.0639 & 1.6484 \\
    \bottomrule
    \end{tabular}
    \end{minipage}
    \hspace{1mm}
    \begin{minipage}[t]{0.145\textwidth}
    \begin{tabular}{|p{0.3cm}|p{0.5cm} p{0.6cm}}
    \toprule
    Prob & Err$_{vel}$ & Err$_{flow}$ \\
    \midrule
    0.1 & 1.0544& 1.6380 \\
    0.7 & 1.0491& 1.6347 \\
    1 & 1.0759 & 1.6526 \\
    \bottomrule
    \end{tabular}
    \end{minipage}
    \hspace{1mm}
    \begin{minipage}[t]{0.15\textwidth}
    \begin{tabular}{|p{0.6cm}|p{0.5cm} p{0.6cm}}
    \toprule
    G/U & Err$_{vel}$ & Err$_{flow}$ \\
    \midrule
    0.3/0.7 & 1.0709 & 1.6476 \\
    0.5/0.5 & 1.0808 & 1.6523 \\
    0.7/0.3 & 1.0669 & 1.6451 \\
    \bottomrule
    \end{tabular}
    \end{minipage}
    \vspace{-3mm}
    \caption{Evaluation with noises (1-3) on Drill$_2$. 1: a zero-mean Gaussian noise with a Std (standard deviation); 2: a 2D uniform noise with Prob being the probability of values within [-0.7, 0.7] $\times$ [-0.8, 0.8]; 3: a mixture noise as the weighted sum of the Gaussian (G) and the uniform (U) noise.}
    \label{tab: nosie_validation}
    \vspace{-5mm}
\end{table}
CCTV camera data is by far the major data on high-density crowd particularly because of the uninvasiveness of the data collection process. However, these videos often suffer from low quality because they are captured from distant cameras and tend to be blurred, making optical flow estimation unreliable. Owning to the physics model in Crowd MPM, our model incorporates a self-correction strategy (P2G operation) that effectively handles spatially noisy. 

Since the noise in optical flows is unknown, we conducted an experiment on Drill$_2$ to show that our method is robust to several noises with various shapes and magnitudes, \eg. Gaussian, uniform, and their mixtures, in \cref{tab: nosie_validation}.  All results are similar, demonstrating the robustness of our model. To understand this, when noisy data are given, the P2G projects it onto the grid (\ie equivalent to one step smoothing) then the mass and momentum are preserved the whole time, essentially filtering out high-frequency noises. During training, even when the ground-truth is noisy, Crowd MPM computes its closest physically sensible prediction, by conserving the mass and momentum. Admittedly, if the optical flow is completely obscure, \eg a white noise, Crowd MPM would not learn the underlying dynamics. In this case, Crowd MPM can still be hand-tuned to visually mimic the data.

\subsection{Visual Analysis Based on Learned Crowd Materials}
\begin{figure}
    \centering
    \includegraphics[width=1\linewidth]{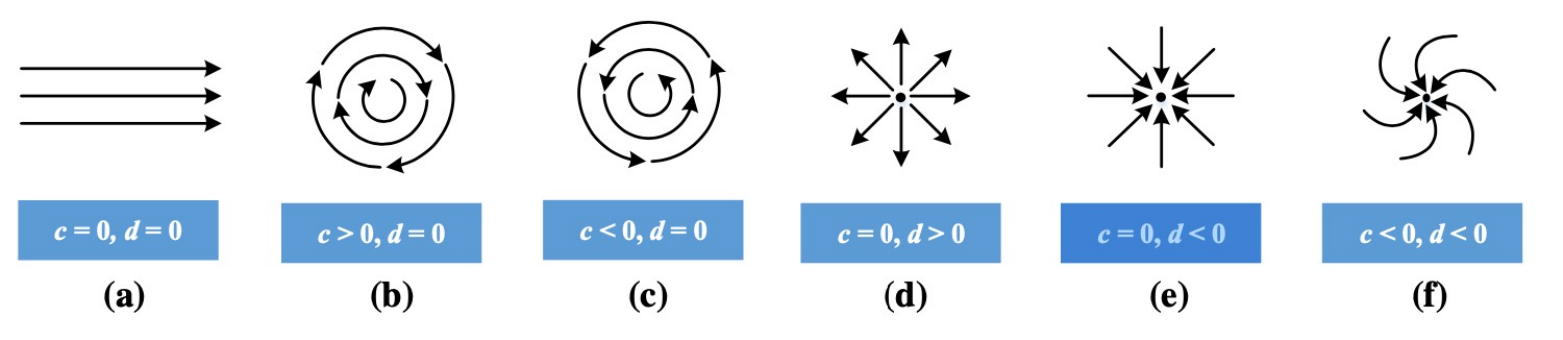}
    \vspace{-6mm}
    \caption{a–e Five simple motion vector fields corresponding to c and d (where c and d denote the value of curl and divergence respectively)\cite{wu2017crowd}.(a) no rotation or dispersal/gathering, (b) the rotation is clockwise and there is no divergence, (c) the rotation is counterclockwise without any dispersal/gathering, (d) the motion is pure dispersal without any curl component, (e) the motion is pure gathering without any curl component, (f) the motion is relatively complex since it has rotation and convergence simultaneously. }
    \label{fig:patterns}
    \vspace{-3mm}
\end{figure}

\begin{figure*}
    \centering
    \begin{subfigure}{1\textwidth}
        \includegraphics[width = 0.95\linewidth]{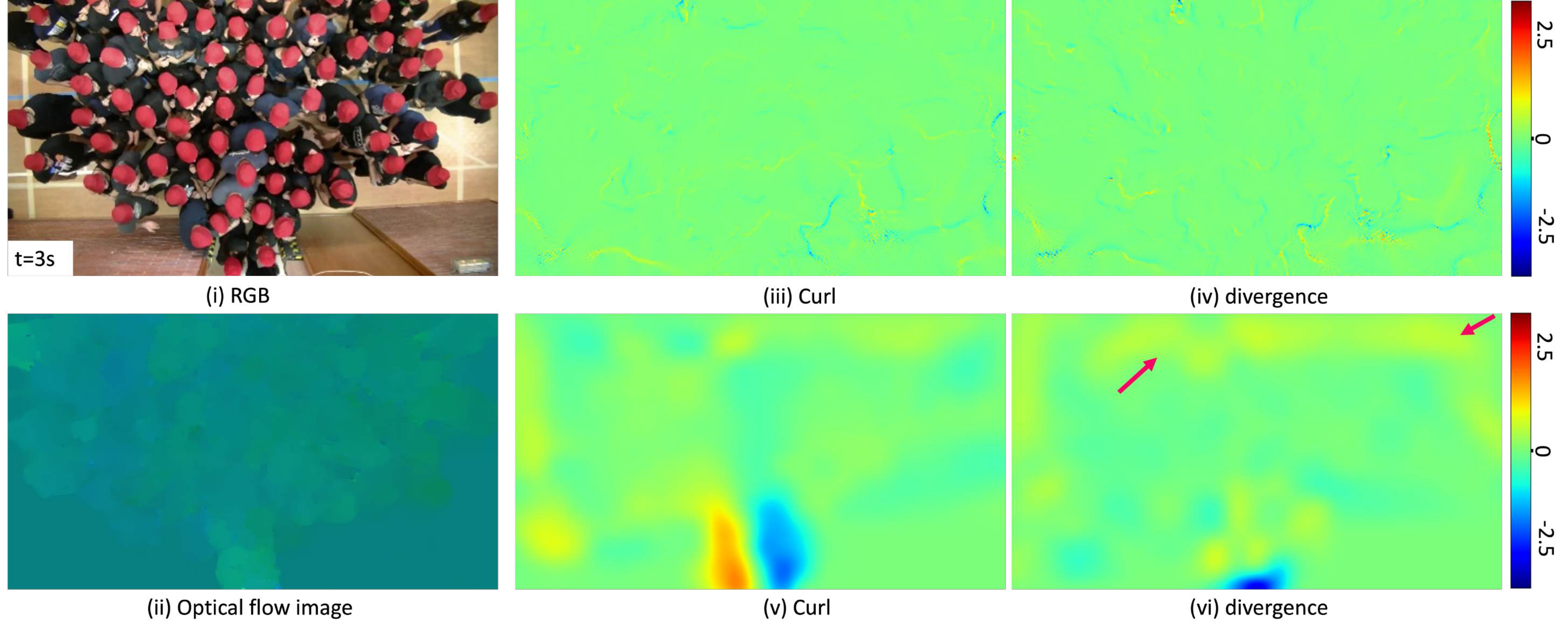}
        \caption{}
        \label{fig:curl_and_div_3}
    \end{subfigure}
    \begin{subfigure}{1\textwidth}
        \includegraphics[width = 0.95\linewidth]{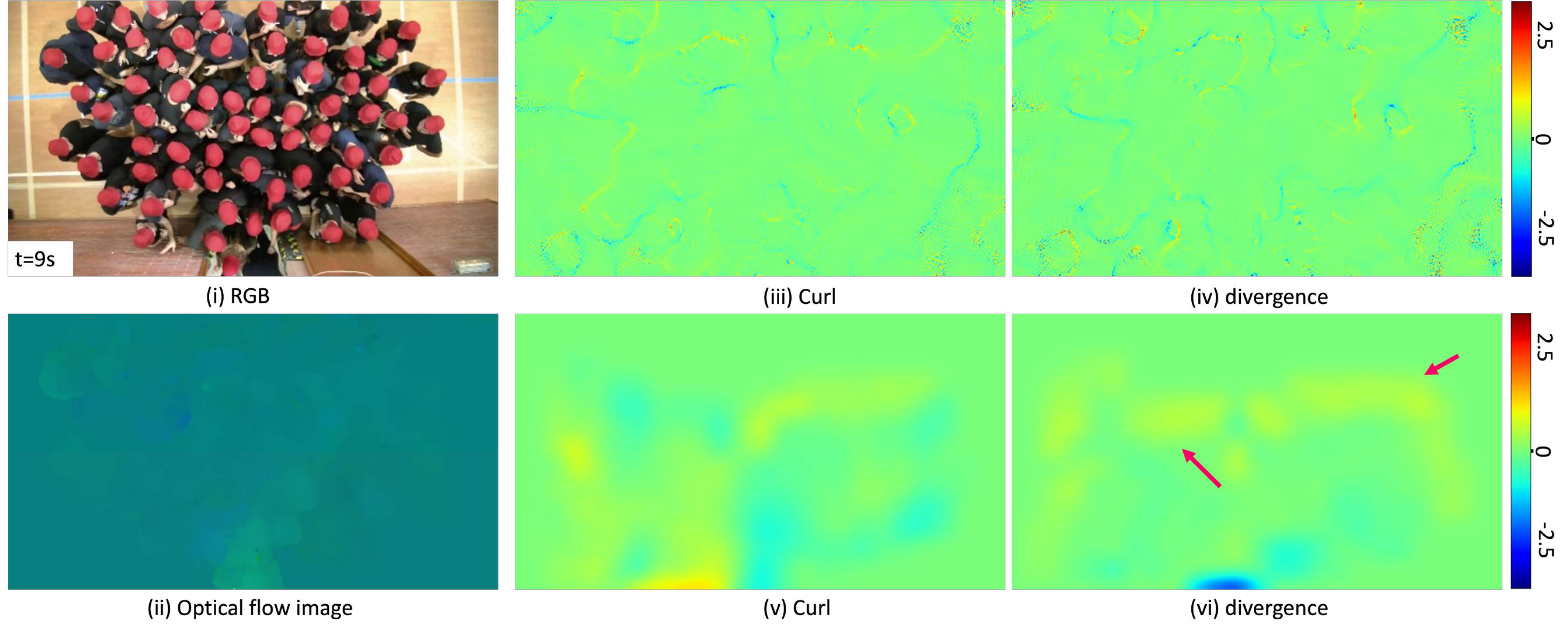}
        \caption{}
        \label{fig:curl_and_div_9}
    \end{subfigure}
    \begin{subfigure}{1\textwidth}
        \includegraphics[width = 0.95\linewidth]{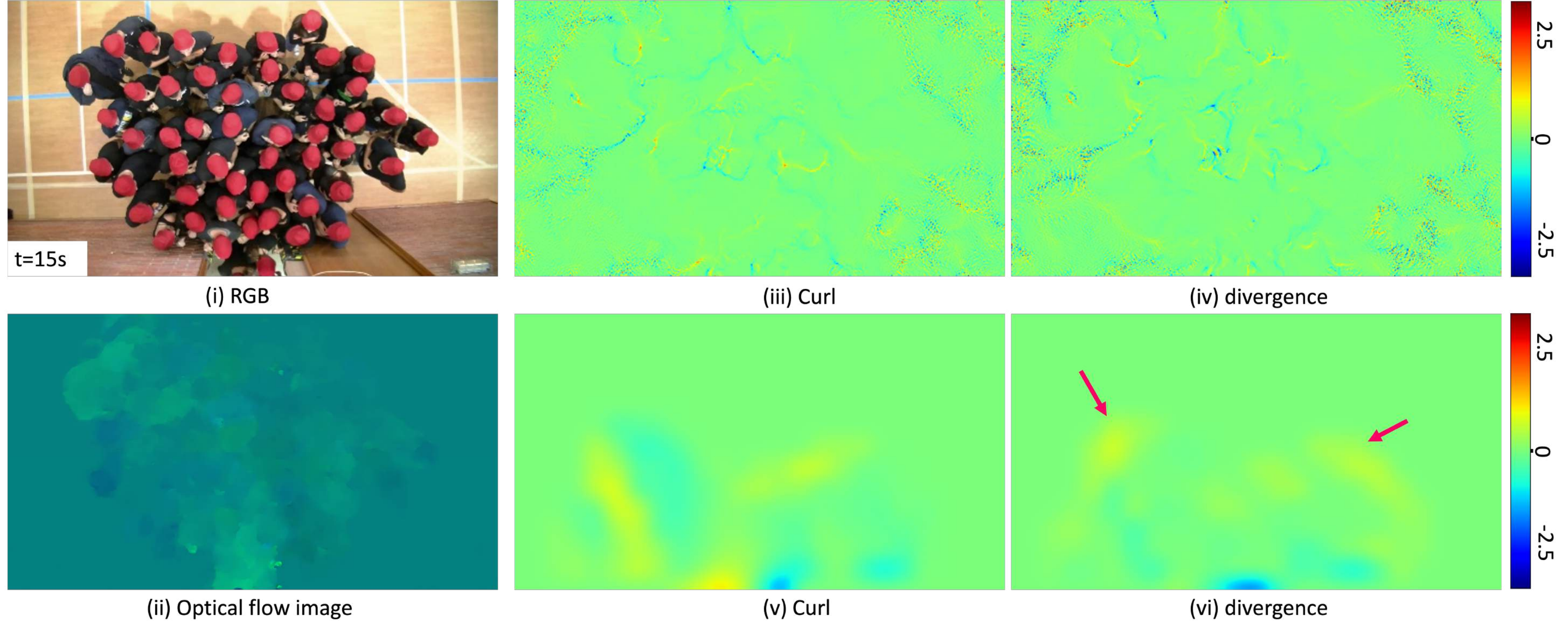}
        \caption{}
        \label{fig:curl_and_div_15}
    \end{subfigure}
    \caption{Qualitative comparison on curl and divergence map. iii and iv are generated from \cite{wu2017crowd}, while v and vi are obtained by our method. Zoom in for a better view.}
    \label{fig:curl_and_divergence}
\end{figure*}
In addition to prediction and simulation, our method is also a good tool for visual analysis of high-density crowds, especially in terms of its global flows. Previously, operators such as curl (vorticity) and divergence have been proposed as a good visualization tool for analysis~\cite{wu2017crowd}.  However, \cite{wu2017crowd} proposes to compute vorticity and divergence on optical flows while our method compute them on the velocity fields. 
Curl is defined as $c = \frac{\partial\vect{v}_y}{\partial x} - \frac{\partial \vect{v}_x}{ \partial y}$, representing the tendency of the flow to exhibit rotational motion or vorticity. Divergence, on the other hand, is defined as $d=\frac{\partial \vect{v}_x}{\partial x} + \frac{\partial \vect{v}_y}{\partial y}$, representing the rate at which a quantity (such as mass or velocity) expands or contracts at a point. Note that $\vect{v}$ is a motion vector field, which can either be optical flow or velocity field.  Several examples of global flow patterns are provided in \cref{fig:patterns}, illustrating how these flows can be analyzed and interpreted using curl and divergence. 

In our scenario, curl reflects steering behaviors, showing how individuals adjust their movements, particularly at the periphery or near obstacles or other peoople. Divergence indicate areas where the crowd density is decreasing (positive divergence) or increasing (negative divergence), such as the region near an exit in Drill where people move towards. We show a qualitative comparison on Drill in \cref{fig:curl_and_divergence} at 3$s$, 9$s$ and 15$s$. 

\cite{wu2017crowd} is designed for crowd behavior classification, such as lane formation, clockwise arch and so on, so it aims to recognize motion patterns based on normalized velocities (per pixel), shown in iii and iv in \cref{fig:curl_and_divergence} (zoom in for a better view). Therefore, the curl and divergence computed based on normalized velocities can only indicate the general patterns of motions, not the actual motions. Visually, the global flow patterns are not obvious. 

In contrast, our model can act as a more detailed analyzer based on the velocity and interpret how the behavior evolves. For example, the areas pointed out by arrows in \cref{fig:curl_and_div_3} vi, \cref{fig:curl_and_div_9} vi and \cref{fig:curl_and_div_15} vi have relatively high positive divergence. Physically, this means there are masses flowing out of these areas. In time, these areas move toward the exit. This captures that the fact that people come into the scene and walk towards the exit. After a short while, there is no people entering the scene. Correspondingly, the high divergence areas first appear near the top entrance then gradually move towards the exit. The divergence in front of the exit is always negative, \ie masses flowing in, due to that for this area the exit rate is lower than the entering rate (\ie people aggregating in front of the exit but not many can get through), hence the blue regions.

In terms of the curl, it reflects the local rotations of masses in a region.  This is reflected by the red and blue areas near the exit at $t=3s$ in \cref{fig:curl_and_div_3} v. This is when the first batch of people arrive at the exit. At this time, people can still go through the exit relatively easily, especially the people right in front of the exit walking relatively faster than the people on both sides of the exit. This caused a relative local rotation of the mass flow, \ie masses in the middle flowing out faster than the ones on both sides, and masses on both sides getting into the middle flow. Both the clockwise (red region) and counter-clockwise (blue region) vorticity reflect this. This curl becomes less prominent at $t=9s$ and $t=12$, when many people are crowded in front of the exit and start to block the way, so that the general movements become slow.

\cref{fig:curl_and_divergence} demonstrates the importance of being able to extract and stably simulate the velocity fields for analysis. Capturing the underlying velocity field provides clear global flow trends, then direct analysis on the optical flows which might be polluted by noises~\cite{wu2017crowd}. Our method provides a tool which not only extracts and simulates the velocity field, but also learns the dynamics from specific crowds.

\subsection{Ablation Study}
We analyze the impact of different components, by evaluating four variants of our model, including different combinations of learnable parameters:
\begin{itemize}
    \item \textbf{\textit{Baseline$_{I}$}}, a weakly incompressible fluid model~\cite{tampubolon2017multi},  with a global, non-learnable parameter $\epsilon$ whose value is obtained by a grid search.
    \item \textbf{\textit{Baseline$_{II}$}} is the same as \textbf{\textit{Baseline$_{I}$}} but with $\epsilon$ learnable for each particle.
    \item \textbf{\textit{Baseline$_{III}$}} has a new strain-stress tensor and new learnable parameter $k$, based on \textbf{\textit{Baseline$_{II}$}}.
    \item \textbf{\textit{Baseline$_{IV}$}} is \textbf{\textit{Baseline$_{III}$}} with $NN_{\alpha}$ (introduced in Sec.3.4 in main paper).
    \item \textbf{\textit{Ours}} is the full model.
\end{itemize}
 \Cref{tab:ablation} clearly shows a continuously improvement across multiple metrics, when more components are added. It has been argued that high-density crowds behave like fluids \cite{treuille2006continuum}. This is confirmed by the hand-tuned \textbf{\textit{Baseline}}$_I$, whose performance is reasonable. \textbf{\textit{Baseline}}$_{II}$ shows the necessity of making the material learnable. With learnable $\epsilon$, the material becomes more heterogeneous and fits the data better. Furthermore, an addition of a learnable compression component (\textbf{\textit{Baseline}}$_{III}$) can capture the dynamics better. We notice that the difference between \textbf{\textit{Baseline}}$_{I-III}$ seems small. After further investigation, this is likely to be caused by two factors. First, \textbf{\textit{Baseline}}$_I$ is a already good baseline as argued by previous research. Hajj crowds form stable flows circling around the center of the scene. Therefore, it is possible to hand-tune \textbf{\textit{Baseline}}$_I$ to mimic the general dynamics, although this relies on heavy human labor. Second, the metrics are based on the optical flow or the velocity field estimated from optical flow, where the scale of values is small.

The results are significantly after active forces are introduced, even with only the motion alignment component in \textbf{\textit{Baseline}}$_{IV}$ with no stochasticity. This further proves such crowds can be interpreted as active matters and learning such active forces greatly enhance the model's ability to predict dynamics. Further adding the stochastic part again significantly improves the results (\textbf{\textit{Ours}}).

\begin{table}[tb]
  \centering
  \begin{tabular}{p{1.6 cm}|p{1.0cm} p{1.0cm} | p{1.0cm} p{1.0cm}}
    \toprule
    \multirow{2}{*}{Methods} & \multicolumn{2}{c|}{Drill$_{1}$} & \multicolumn{2}{c}{Hajj}  \\
    \cline{2-5}
     & Err$_{vel}$ & Err$_{flow}$ & Err$_{vel}$ & Err$_{flow}$ \\
    \midrule
     \textbf{\textit{Baseline$_{I}$}}  & 0.60740 & 0.98405 & 1.45878 & 1.68066 \\
     \textbf{\textit{Baseline$_{II}$}}  & 0.60727 & 0.98396 & 1.45709 & 1.67930 \\ 
     \textbf{\textit{Baseline$_{III}$}}  & 0.60483 & 0.98160  & 1.45679 & 1.67906\\ 
     \textbf{\textit{Baseline$_{IV}$}}  & 0.48987 & 0.88263 & 0.76161 & 1.07112 \\
     \midrule
     \textbf{\textit{Ours(mean)}} & 0.48276 & 0.87165 & 0.69135 & 0.99511 \\
     \textbf{\textit{Ours(best)}} & 0.48161 & 0.87033 & 0.68281 & 0.98604 \\
    \bottomrule
  \end{tabular}
  \caption{Ablation study on $Drill_1$ and $Hajj$. Results involving active force are based on 10 trials. The training and testing duration are 60 frames on $Drill_1$ and 30 frames on $Hajj$.}
  \label{tab:ablation}
  \vspace{-0.6cm}
\end{table}

\section{Crowd Material Point Method}
There are two mainstream methods to solve PDEs. One is the Lagrangian method which treats the material as a set of particles and tracks them to determine their properties such as mass, positions, and velocities. The other one is the Eulerian method which discretizes the space into grids containing nodes with various properties and employs the grids to represent materials. Material Point Method (MPM) is a hybrid Eulerian-Lagrangian method combining the advantages of Eulerian and Lagrangian methods. As our model involves Eulerian data and Lagrangian behaviors, we enhance the standard MPM to propose Crowd MPM to simulate extremely high-density crowds.

Our derivations are mainly following \cite{jiang2016material}. The main changes are brought by our newly introduced learnable stress and active forces. Similar to the standard MPM, our Crowd MPM also incorporates both the Lagrangian view and the Eulerian view. Under the Lagrangian view, we focus on the particle $p$ with mass $m_p$, position $\vect{x}_p$, and velocity $\vect{v}_p$. The motion of material is defined by a deformation map $\Phi(\cdot, t): \Omega^0 \rightarrow \Omega^t$, where $\Omega^0,  \Omega^t \subset \mathbb{R}^2$. $\Omega^t$ denotes the set of positions of particles at time $t$. As a result, $\Omega^0$ is the set of initial positions of particles. Particularly, we use $\vect{X}_p$ and $\vect{x}_p$ to denote the initial position and the current position for any particle in the simulated material, respectively. We can note that $\vect{X}_p = \vect{x}_p^0$. The deformation map $\Phi$ further enables us to determine the velocity (we ignore the subscript $p$ for notation simplicity in this section):
\begin{equation}
    \vect{V}(\vect{X},t) = \frac{\partial \Phi}{\partial t}(\vect{X},t).
\end{equation}
The Jacobian of the deformation map $\Phi$ is significantly important and will be used often later:
\begin{equation}
    \vect{F}(\vect{X},t)= \frac{\partial \Phi}{\partial \vect{X}}(\vect{X},t).
    \label{eq:defMt}
\end{equation}
The most common way to employ $\vect{F}$ is to use its determinant $J$. Finally, we give the governing equations based on the conservation of mass and conservation of momentum:
\begin{equation}
    R(\vect{X},t)J(\vect{X},t)=R(\vect{X},0),
    \label{eq:sm_lcm}
\end{equation}
\begin{equation}
    R(\vect{X},0)\frac{\partial \vect{V}}{\partial t} = \nabla^{\vect{X}} \cdot \vect{P^{cm}} + R(\vect{X},0)(\vect{B^{bd}}+ \vect{B^{act}}),
    \label{eq:sm_lcmv}
\end{equation}
where $R$ is the Lagrangian mass density, $\vect{P^{cm}}$ is the crowd material stress under the Lagrangian view, and $\vect{B^{bd}}$ and $\vect{B^{act}}$ denote the accelerations deriving from the body force and the active force on $\vect{X}$, respectively. To be more specific, the Lagrangian mass density is defined via:
\begin{align}
 R(\vect{X},t)&=\rho(\Phi(\vect{X},t),t)=\rho(\vect{x},t), \\
 \rho(\vect{x},t) &= \lim_{\epsilon\rightarrow+0} \frac{mass(B_{\epsilon}^t)}{\int_{B_{\epsilon}^t}d\vect{x}},
\end{align}
where $B_{\epsilon}^t \subset \Omega^t$ is a ball with radius $\epsilon$ and center $x \in \Omega^t$. We note that the Lagrangian view builds governing equations on material space $\Omega^0$. We offer the proof for \cref{eq:sm_lcm} and \cref{eq:sm_lcmv} in \cref{sec:sm_proof}. 

Under the Eulerian view, we pay attention to the world space $\Omega^t$. In addition, grids consisting of a series of nodes with index $\textbf{i}$ are built to estimate material motion. Each note $\textbf{i}$ has physics properties including mass $m_{\textbf{i}}$, position $\vect{x}_{\textbf{i}}$, and velocity $\vect{v}_{\textbf{i}}$, \etc. We also give the governing equations based on the conservation of mass and conservation of momentum:
\begin{equation}
    \frac{D}{Dt}\rho(\vect{x}, t) + \rho(\vect{x}, t)\nabla^{\vect{x}} \cdot \vect{v}(\vect{x}, t) = 0
    \label{eq:sm_ecm}
\end{equation}
\begin{equation}
    \rho(\vect{x}, t)\frac{D\vect{v}}{Dt}=\nabla^{\vect{x}}\cdot\vect{\sigma^{cm}}+\rho(\vect{x}, t)(\vect{b^{bd}}+ \vect{b^{act}})
    \label{eq:sm_ecmv}
\end{equation}
where $\frac{D}{Dt}$ is the material derivative~\cite{temam2001navier}, $\vect{v}(\vect{x}, t)=\vect{V}(\Phi^{-1}(\vect{x}, t),t)$, $\vect{\sigma^{cm}}$ is the crowd material stress under the Eulerian view, and $\vect{b^{bd}}$ and $\vect{b^{act}}$ denote the accelerations deriving from the body force and the active force on $\vect{x}$, respectively. We also leave the proof for \cref{eq:sm_ecm} and \cref{eq:sm_ecmv} in \cref{sec:sm_proof}

Finally, we give the weak form, which is crucial for the discretization, of the conservation of momentum in the Eulerian view: 
\begin{align}
    \int_{\Omega^t} q_i \rho a_i d\vect{x} = \int_{\partial \Omega^t} q_i t_i ds(\vect{x}) - \int_{\Omega^t} q_{i, k} \sigma_{ik}^{cm}d\vect{x} \notag \\
    +\int_{\Omega^t} q_i \rho b^{bd}_i d\vect{x} + \int_{\Omega^t} q_i \rho b^{act}_i d\vect{x}.
    \label{eq:sm_wfe}
\end{align}
where $q_i$ is the ith component of an arbitrary function $\vect{q}(\cdot,t):\Omega^t \rightarrow \mathbb{R}^d$, $q_{i, k} = \frac{\partial q_i}{\partial x_k}$, $a_i$ is the ith component of $\boldsymbol{a}=\frac{D\vect{v}}{Dt}$, and $t_i$ is the ith component of the boundary force per unit reference area $\vect{t}(\vect{x},t)$. Similarly, $\sigma_{ik}^{cm}$, $b^{bd}_i$ and $b^{act}_i$ are the components of $\vect{\sigma^{cm}}$, $\vect{b^{bd}}$, and $\vect{b^{act}}$, respectively. $ds$ denotes a tiny area. For efficient expression, the summation is implied on the repeated index. The proof of the weak form can be found in \cref{sec:sm_proof}.

Subsequently, we can clarify three steps in our Crowd MPM individually: (1) Particle-to-grid transfer in \cref{sec:sm_1_1} (2) Grid Operation in \cref{sec:sm_1_2}, and (3) Grid-to-particle transfer in \cref{sec:sm_1_3}.

\subsection{Particle-to-grid Transfer}
\label{sec:sm_1_1}
This step aims to transfer particle mass and momentum to the grid at each timestamp $n$. Each particle contributes to the transfer process based on weights relying on particles and grids. Specifically, the transfer process in defined by:    
\begin{equation}
    m_{\textbf{i}}^n = \int_{\vect{x}\in \Omega^n} W_{\textbf{i}}^n(\vect{x})\rho(\vect{x})\ d\vect{x},
    \label{eq:sm_p2g_i_m}
\end{equation}
\begin{equation}
m_{\textbf{i}}^n\vect{v}_{\textbf{i}}^n=\int_{\vect{x}\in \Omega^n}W_{\textbf{i}}^{n'}(\vect{x})\rho(\vect{x})\vect{v}(\vect{x}) d\vect{x}, 
    \label{eq:sm_p2g_i_mv}
\end{equation}

where $W_{\textbf{i}}^n$ and $W_{\textbf{i}}^{n'}$ are weight functions, $\Omega^n$ is the set of all particles at timestamp $n$, $\rho$ denotes density.

We follow the Affine Particle-In-Cell (APIC) method ~\cite{jiang2015affine} to discretize \cref{eq:sm_p2g_i_m} and \cref{eq:sm_p2g_i_mv} due to its excellent numerical properties:
\begin{align}
&m_{\textbf{i}}^n = \sum_p w_{\textbf{i}p}^nm_p, \\
&m_{\textbf{i}}^n\vect{v}_{\textbf{i}}^n = \sum_p w_{\textbf{i}p}^n[m_p\vect{v}_p^n + m_p\mat{C}_p^n(\vect{x}_{\textbf{i}}-\vect{x}_p^n)],
\end{align}
where $w_{\textbf{i}p}^n = \phi(\vect{x}_{\textbf{i}} - \vect{x}_p^n)$ ($\phi$ is a quadratic B-spline function~\cite{hu2018moving}), $\mat{C}_p^n$ is the affine velocity gradient ~\cite{jiang2015affine}. $m_p$ and $\vect{x}_{\textbf{i}}$ don't have superscript $n$ because $m_p$ doesn't change with time and the grids are rebuilt at each MPM iteration.

\subsection{Grid Operation}
\label{sec:sm_1_2}

We update velocities on the grid by:
\begin{equation}
    \vect{v}_{\textbf{i}}^{n+1} = \vect{v}_{\textbf{i}}^{n} + \Delta t \frac{\vect{f}_{\textbf{i}}^n}{m_{\textbf{i}}}, 
\label{eq:sm_gridUpdate}
\end{equation}
Then boundary conditions are employed to refine the updated velocities:
\begin{equation}
    \vect{v}_{\textbf{i}}^{n+1} = \textbf{BC}(\vect{v}_{\textbf{i}}^{n+1}) = \vect{v}_{\textbf{i}}^{n+1} - \gamma\vect{n}\left< \vect{n},\vect{v}_{\textbf{i}}^{n+1}\right>.
\end{equation}
We explain how to obtain $\vect{v}_{\textbf{i}}^{n}$ and $m_{\textbf{i}}$ in the last section. Next, I will introduce how to estimate $\vect{f}_{\textbf{i}}^n$. 

Like \cref{sec:sm_1_1}, we transfer forces on particles to the grid at each timestamp $n$. We give the transfer process for each component of forces. To avoid confusion, we use Greek letters like $\alpha$ and $\beta$ to denote the component index. 
\begin{align}
\vect{f}_{\textbf{i}\alpha}^n=&\int_{\vect{x}\in \Omega^n}
   W_{\textbf{i}}^n(\vect{x}) \rho(\vect{x}, t)\boldsymbol{a}_{\alpha}(\vect{x}, t)d\vect{x} 
\end{align}
where other components of $\vect{f}_{\textbf{i}}^n$ has a similar transfer process. According to the weak form of the conservation of momentum in the Eulerian view \cref{eq:sm_wfe}, we have:
\begin{align}
\vect{f}_{\textbf{i}\alpha}=\int_{\Omega^t} W_{\textbf{i}} \rho a_{\alpha} d\vect{x} &= \int_{\partial \Omega^t} W_{\textbf{i}} t_{\alpha} ds(\vect{x}) - \int_{\Omega^t} W_{\textbf{i}, \gamma} \sigma_{\alpha \gamma}^{cm}d\vect{x} \notag \\
    &+\int_{\Omega^t} W_{\textbf{i}} \rho b^{bd}_{\alpha} d\vect{x} + \int_{\Omega^t} W_{\textbf{i}} \rho b^{act}_{\alpha} d\vect{x}.
    \label{eq:sm_fc}
\end{align}
where the superscript is ignored for notation simplicity. Here, we set $q_{\alpha} = W_{\textbf{i}}$ and $q_{\gamma} = 0$ because $\vect{q}$ is arbitrary. Following previous work ~\cite{jiang2016material}, we omit the $\int_{\partial \Omega^t} W_{\textbf{i}} t_{\alpha} ds(\vect{x})$ in \cref{eq:sm_fc} for simplicity given its minor effect. Given our estimation of $\vect{\sigma_{p}^{cm}}=\sigma(\vect{x}_p,t)$ at every Lagrangian particle $\vect{x}_p$, we have:
\begin{align}
f_{\textbf{i}\alpha}^{st} = -\int_{\Omega^t} W_{\textbf{i}, \gamma} \sigma_{\alpha \gamma}^{cm}d\vect{x} &\approx -\sum_p \sigma_{p \alpha \gamma}^{cm} w_{\textbf{i}p,\gamma} V_p  \notag \\
&= -\sum_p \vect{\sigma_{p \alpha}^{cm}} \nabla w_{\textbf{i}p} V_p,
\label{eq:sm_fst_d1}
\end{align}
where $\vect{\sigma_{p \alpha}^{cm}}$ is the $\alpha$th row of $\vect{\sigma_{p}^{cm}}$, $V_p$ is the volume of $B_{\Delta x}^t$ at the Lagrangian particle $\vect{x}_p$. We follow ~\cite{jiang2015affine} to estimate $V_p$. Because we use the quadratic B-spline function to represent $w_{\textbf{i}p}$, we can obtain its gradient by:
\begin{align}
    \nabla w_{\textbf{i}p} = \frac{4}{\Delta x^2}w_{\textbf{i}p}(\vect{x}_\textbf{i}-\vect{x}_p)
    \label{eq:sm_fst_d2}
\end{align}
Combining \cref{eq:sm_fst_d1} and \cref{eq:sm_fst_d2} results in:
\begin{align}
    f_{\textbf{i}\alpha}^{st} \approx -\sum_p  \frac{4}{\Delta x^2}w_{\textbf{i}p} \vect{\sigma_{p \alpha}^{cm}} (\vect{x}_\textbf{i}-\vect{x}_p)  V_p
    \label{eq:sm_fst}
\end{align}
By defining:
\begin{align}
&\frac{4}{\Delta x^2} \vect{\sigma_{p \alpha}^{cm}} (\vect{x}_\textbf{i}-\vect{x}_p)  V_p
    = \notag \\ &\mat{G}_p(\vect{x}_i-\vect{x}_p) + \sum_{p'\in N_p}(\func{f}_{r}^{pp'} - \func{f}_{t}^{p p'}),
\end{align}
we can get the Eq.11 in the main paper. For the body force, we take the centripetal force as an example. Other body forces such as the goal attraction force have a similar derivation. Therefore, $\vect{b^{bd}}(\vect{x}, t)=\frac{\vect{v}(\vect{x}, t)^2}{r}$, where $r$ is the length of the radius of the moving circle. Then, we have:
\begin{align}
    f^{bd}_{\textbf{i}\alpha}=\int_{\Omega^t} W_{\textbf{i}} \rho b^{bd}_{\alpha} d\vect{x} &= \int_{\Omega^t} W_{\textbf{i}} \rho \frac{\vect{v}_{\alpha}^2}{r} d\vect{x} \notag \\ 
    &\approx \sum_p w_{\textbf{i}p} m_p \frac{\vect{v}_{p \alpha}^2}{r}.
        \label{eq:sm_fbd}
\end{align}
As for the active force, we have $\vect{b^{act}}=a\vect{v} + \vect{c}$, where $\vect{c}$ is the stochastic part of the active force, \ie $\vect{c} = - b|\vect{v}|^2\vect{v} + D_L\nabla(\nabla\cdot\vect{v}) +D_1\nabla^2\vect{v} + D_2(\vect{v}\cdot\nabla)^2\vect{v} + \Tilde{\vect{f}}$. $a$ and $\vect{c}$ are predicted by neural networks. Then, we have:
\begin{align}
    f^{act}_{\textbf{i}\alpha}=\int_{\Omega^t} W_{\textbf{i}} \rho b^{act}_{\alpha} d\vect{x} &= \int_{\Omega^t} W_{\textbf{i}} \rho ( a\vect{v}_{\alpha} + c_{\alpha}) d\vect{x} \notag \\ 
    &= \sum_p w_{\textbf{i}p} m_p (a\vect{v}_{p \alpha} + c_{p \alpha}).
    \label{eq:sm_fact}
\end{align}
Combining \cref{eq:sm_fst}, \cref{eq:sm_fbd} and \cref{eq:sm_fact}, we have:
\begin{align}
    \vect{f}_{\textbf{i}\alpha} = f^{st}_{\textbf{i}\alpha} + f^{bd}_{\textbf{i}\alpha}+ f^{act}_{\textbf{i}\alpha} \approx \sum_p  -\frac{4}{\Delta x^2}w_{\textbf{i}p} \vect{\sigma_{p \alpha}^{cm}} (\vect{x}_\textbf{i}-\vect{x}_p)  V_p \notag \\
    + w_{\textbf{i}p} m_p \frac{\vect{v}_{p \alpha}^2}{r} + w_{\textbf{i}p} m_p (a\vect{v}_{p \alpha} + c_{p \alpha})
\end{align}

\subsection{Grid-to particle Transfer}
\label{sec:sm_1_3}
Finally, we transfer the updated velocities $\vect{v}_{\textbf{i}}^{n+1}$ back to particles and update positions of these particles:
\begin{align}
    \vect{v}_p^{n+1} = \sum_{\textbf{i}} w_{\textbf{i}p}^n\vect{v}_{\textbf{i}}^{n+1}, \,\, 
    \vect{x}_p^{n+1} = \vect{x}_p^n + \Delta t \vect{v}_p^{n+1},
\end{align}
We update the velocity gradient $\vect{C}_p^{n+1}$ for next step:
\begin{align}
    \mat{C}_p^{n+1} = \frac{4}{\Delta x^2}\sum_i w_{\textbf{i}p}^{n} \vect{v}_{\textbf{i}}^{n+1}(\vect{x}_{\textbf{i}}-\vect{x}_p^{n})^{T}.
\end{align}
Additionally, we also update the deformation gradient $\vect{F}_p^{n+1}$ for estimating $V_p$:
\begin{align}
    \mat{F}_p^{n+1} =(\mat{I} + \Delta t \mat{C}_p^{n+1})\mat{F}_p^{n}.
\end{align}

\section{Proof Details}
\label{sec:sm_proof}
\textbf{Push Forward and Pull Back.} We introduce the push forward and pull back operations based on the deformation map $\Phi$ for efficient formula derivation and understanding. The deformation map $\Phi:\Omega^0 \rightarrow \Omega^t$ typically is assumed to be bijective, which means that any function defined on one set like $\Omega^0$ can naturally be regarded as another function defined on another set like $\Omega^t$ by changing variables. Push forward and pull back are two ways of variable substitution here. To be formal, given the time $t$ and a function $G(\cdot,t):\Omega^0 \rightarrow \mathbb{R}$, we define the push forward $g(\cdot,t):\Omega^t \rightarrow \mathbb{R}$ as $g(\vect{x},t)=G(\Phi^{-1}(\vect{x},t),t)$ with $\vect{x} \in \Omega^t $. Similarly, given the time t and a function $g(\cdot,t):\Omega^t \rightarrow \mathbb{R}$, the pull back $G(\cdot,t):\Omega^0 \rightarrow \mathbb{R}$ is defined as $G(\vect{X},t)=g(\Phi(\vect{X},t),t)$ with $\vect{X} \in \Omega^0$. Particularly, the pull back of the pull forward of a function such as $G(\cdot,t)$ is itself. We commonly think of a function of $\vect{X}$/$\vect{x}$ as Lagrangian/Eulerian. Therefore, the push forward (pull back) of a function which is Lagrangian (Eulerian) is Eulerian (Lagrangian).            

\textbf{Conservation of Mass under the Lagrangian View.} For an arbitrary $\vect{x} \in \Omega^t$, we consider the ball $B_{\epsilon}^t \subset \Omega^t$ with radius $\epsilon$ and center $\vect{x}$. The mass in $B_{\epsilon}^t$ shouldn't vary over time when $\epsilon$ is small enough. Therefore, we have $mass(B_{\epsilon}^t) = mass(B_{\epsilon}^0)$. According to the definition of $\rho$, $R$, and $J$, we have:
\begin{equation}
    mass(B_{\epsilon}^t) = \int_{B_{\epsilon}^t} \rho(\vect{x}, t)d\vect{x} = \int_{B_{\epsilon}^0} R(\vect{X}, t)Jd\vect{X},
    \label{eq:sm_mbt}
\end{equation}
\begin{equation}
    mass(B_{\epsilon}^0) = \int_{B_{\epsilon}^0} R(\vect{X}, 0)d\vect{X},
    \label{eq:sm_mb0}
\end{equation}
for all $B_{\epsilon}^t \subset \Omega^t$ ($t\geq0$), where the second equality in \cref{eq:sm_mbt} is obtained by the formula of changing variables. From \cref{eq:sm_mbt} and \cref{eq:sm_mb0}, we can derive:
\begin{equation}
    \int_{B_{\epsilon}^0} R(\vect{X}, t)Jd\vect{X} = \int_{B_{\epsilon}^0} R(\vect{X}, 0)d\vect{X}.
\end{equation}
Given that $B_{\epsilon}^t$ is arbitrary, we can obtain:
\begin{equation}
     R(\vect{X}, t)J(\vect{X}, t) = R(\vect{X}, 0), 
     \label{eq:sm_lcm2}
\end{equation}
for any  $\vect{X} \in \Omega^0$.

\textbf{Conservation of Mass under the Eulerian View.} We start from \cref{eq:sm_lcm2}, \ie \cref{eq:sm_lcm}, to derive the conservation of mass equation under the Eulerian view. First, we have:
\begin{equation}
    \frac{\partial}{\partial t} ( R(\vect{X}, t)J(\vect{X}, t)) =  \frac{\partial}{\partial t} R(\vect{X}, 0) = 0,
\end{equation}
according to \cref{eq:sm_lcm2}. Then, it is easy to get:
\begin{equation}
    \frac{\partial}{\partial t}(RJ) = \frac{\partial R}{\partial t} J + R\frac{\partial J}{\partial t}=0,
    \label{eq:sm_ecm_rjt}
\end{equation}
where we neglect $(\vect{X}, t)$ for notation simplicity. Then, we use the result from ~\cite{jiang2016material}:
\begin{equation}
    \frac{\partial J}{\partial t} = J  \frac{\partial v_1}{\partial x_1} +  J \frac{\partial v_2}{\partial x_2},
    \label{eq:sm_ecm_jt}
\end{equation}
where $\vect{x} = (x_1, x_2) \in \Omega^t$ and $\vect{v}(\vect{x},t) = (v_1, v_2)$. Combining \cref{eq:sm_ecm_rjt}
and \cref{eq:sm_ecm_jt} results in:
\begin{equation}
    \frac{\partial R}{\partial t} J + RJ(  \frac{\partial v_1}{\partial x_1} +   \frac{\partial v_2}{\partial x_2})=0.
    \label{eq:sm_ecm_rjt2}
\end{equation}
We push forward on both sides of \cref{eq:sm_ecm_rjt2} to obtain:
\begin{equation}
    \frac{D}{Dt}\rho(\vect{x}, t) + \rho(\vect{x}, t)\nabla^{\vect{x}} \cdot \vect{v}(\vect{x}, t) = 0.
\end{equation}

\textbf{Conservation of Momentum under the Eulerian View.} Given an arbitrary $B_{\epsilon}^t \subset \Omega^t$, the momentum change on $B_{\epsilon}^t$ can be expressed as:
\begin{align}
     \frac{d}{dt}\int_{B_{\epsilon}^t} \rho(\vect{x}, t) \vect{v}(\vect{x}, t)d\vect{x} = \int_{\partial B_{\epsilon}^t} \vect{\sigma^{cm}}\vect{n}ds(\vect{x}) + \notag \\ 
     \int_{B_{\epsilon}^t} \rho(\vect{x}, t) (\vect{b^{bd}}+ \vect{b^{act}}) d\vect{x},
     \label{eq:sm_cmve_d}
\end{align}
where $\vect{n}(\vect{x})$ is the unit outward normal of $\partial B_{\epsilon}^t$ at $\vect{x}$ and $ds$ denotes a tiny area. We further have:
\begin{align}
     \frac{d}{dt}\int_{B_{\epsilon}^t} \rho(\vect{x}, t) &\vect{v}(\vect{x}, t)d\vect{x} =  \frac{d}{dt}\int_{B_{\epsilon}^0} R(\vect{X}, t) \vect{V}(\vect{X}, t)Jd\vect{X} \notag\\
     &= \int_{B_{\epsilon}^0} R(\vect{X}, t)J(\vect{X}, t) \vect{A}(\vect{X}, t)d\vect{X}, 
    \label{eq:sm_cmve_d2}
\end{align}
where $\vect{A}(\vect{X}, t) = \frac{\partial^2 \Phi}{\partial t^2}(\vect{X},t)=\frac{\partial \vect{V}}{\partial t}(\vect{X},t)$, the first equality comes from the formula of changing variables, and the second equality obtained by swapping the order of the taking derivative symbol and the integral symbol. Combining \cref{eq:sm_cmve_d} and \cref{eq:sm_cmve_d2} results in:
\begin{align}
    &\int_{B_{\epsilon}^0} R(\vect{X}, t)J(\vect{X}, t) \vect{A}(\vect{X}, t)d\vect{X} = \notag\\
     &\int_{\partial B_{\epsilon}^t} \vect{\sigma^{cm}}\vect{n}ds(\vect{x}) + \int_{B_{\epsilon}^t} \rho(\vect{x}, t) (\vect{b^{bd}}+ \vect{b^{act}}) d\vect{x}
     \label{eq:sm_ecm_rja}
\end{align}
We push forward the left side of \cref{eq:sm_ecm_rja} to get:
\begin{equation}
   \int_{B_{\epsilon}^0} R(\vect{X}, t)J(\vect{X}, t) \vect{A}(\vect{X}, t)d\vect{X} = \int_{B_{\epsilon}^t} \rho(\vect{x}, t) \vect{a}(\vect{x}, t)d\vect{x},   
\end{equation}
where $\vect{a}(\vect{x}, t) = \vect{A}(\Phi^{-1}(\vect{x}, t), t)$. Further, we also have $\vect{A}(\Phi^{-1}(\vect{x}, t), t) = \frac{D\vect{v}}{Dt}$, which has proof in ~\cite{jiang2016material}. Then, \cref{eq:sm_ecm_rja} becomes:
\begin{align}
    &\int_{B_{\epsilon}^t} \rho(\vect{x}, t) \vect{a}(\vect{x}, t)d\vect{x} = \int_{B_{\epsilon}^t} \rho(\vect{x}, t) \frac{D\vect{v}}{Dt}d\vect{x} = \notag \\
    &\int_{\partial B_{\epsilon}^t} \vect{\sigma^{cm}}\vect{n}ds(\vect{x}) + \int_{B_{\epsilon}^t} \rho(\vect{x}, t) (\vect{b^{bd}}+ \vect{b^{act}}) d\vect{x} = \notag \\
    &\int_{B_{\epsilon}^t} \nabla^{\vect{x}} \cdot \vect{\sigma^{cm}} d\vect{x} + \int_{B_{\epsilon}^t} \rho(\vect{x}, t) (\vect{b^{bd}}+ \vect{b^{act}}) d\vect{x}
\end{align}
Since $B_{\epsilon}^t$ is arbitrary, we have:
\begin{equation}
    \rho(\vect{x}, t)\frac{D\vect{v}}{Dt}=\nabla^{\vect{x}}\cdot\vect{\sigma^{cm}}+\rho(\vect{x}, t)(\vect{b^{bd}}+ \vect{b^{act}}).
\end{equation}

\textbf{Conservation of Momentum under the Lagrangian View.} We can also derive the conservation of momentum under the Lagrangian view from \cref{eq:sm_ecm_rja}. Instead of pushing forward, we pull back the right side of \cref{eq:sm_ecm_rja}. The pull back of the first integral on the right side of \cref{eq:sm_ecm_rja} is:
\begin{align}
    &\int_{\partial B_{\epsilon}^t} \vect{\sigma^{cm}}(\vect{x}, t) \vect{n}ds(\vect{x}) = \notag \\
    &\int_{\partial B_{\epsilon}^0} J(\vect{X}, t) \vect{\sigma^{cm}}(\Phi(\vect{X}, t), t)\vect{F}^{-T}(\vect{X}, t)\vect{N}(\vect{X})ds(\vect{X}),
\end{align}
where $\vect{N}(\vect{X})$ is the unit outward normal of $\partial B_{\epsilon}^0$ at $\vect{X}$. We introduce our the Crowd first Piola Kirchoff stress $\vect{P^{cm}}=J\vect{\sigma^{cm}}\vect{F}^{-T}$ and get:
\begin{align}
    \int_{\partial B_{\epsilon}^t} \vect{\sigma^{cm}}(\vect{x}, t) \vect{n}ds(\vect{x}) &= \int_{\partial B_{\epsilon}^0} \vect{P^{cm}}(\vect{X}, t)\vect{N}ds(\vect{X}) \notag \\
    & = \int_{B_{\epsilon}^0} \nabla^{\vect{x}} \cdot \vect{P^{cm}} (\vect{X}, t) d\vect{X}.
    \label{eq:sm_cmvl_d1}
\end{align}
Subsequently, we pull back the second integral on the right side of \cref{eq:sm_ecm_rja}:
\begin{align}
    &\int_{B_{\epsilon}^t} \rho(\vect{x}, t) (\vect{b^{bd}}+ \vect{b^{act}}) d\vect{x} = \notag \\ &\int_{B_{\epsilon}^0} R(\vect{X}, t)J(\vect{X}, t) (\vect{B^{bd}}+ \vect{B^{act}}) d\vect{X}= \notag \\
     &\int_{B_{\epsilon}^0} R(\vect{X}, 0) (\vect{B^{bd}}+ \vect{B^{act}}) d\vect{X},
     \label{eq:sm_cmvl_d2}
\end{align}
where we use \cref{{eq:sm_lcm2}} to get the second equality. We note that the left side of \cref{eq:sm_ecm_rja} can be written as using \cref{{eq:sm_lcm2}}:
\begin{align}
    &\int_{B_{\epsilon}^0} R(\vect{X}, t)J(\vect{X}, t) \vect{A}(\vect{X}, t)d\vect{X} = \notag \\   &\int_{B_{\epsilon}^0} R(\vect{X}, 0) \vect{A}(\vect{X}, t)d\vect{X} = \int_{B_{\epsilon}^0} R(\vect{X}, 0) \frac{\partial \vect{V}}{\partial t}(\vect{X},t)d\vect{X}.
    \label{eq:sm_cmvl_d3}
\end{align}
Combining \cref{eq:sm_ecm_rja}, \cref{eq:sm_cmvl_d1}, \cref{eq:sm_cmvl_d2}, and \cref{eq:sm_cmvl_d3} results in:
\begin{align}
    &\int_{B_{\epsilon}^0} R(\vect{X}, 0) \frac{\partial \vect{V}}{\partial t}(\vect{X},t)d\vect{X} = \notag \\
    &\int_{B_{\epsilon}^0} \nabla^{\vect{x}} \cdot \vect{P^{cm}} (\vect{X}, t) d\vect{X} + \int_{B_{\epsilon}^0} R(\vect{X}, 0) (\vect{B^{bd}}+ \vect{B^{act}}) d\vect{X}.
\end{align}
Because $B_{\epsilon}^0$ is arbitrary, we can derive:
\begin{equation}
    R(\vect{X},0)\frac{\partial \vect{V}}{\partial t} = \nabla^{\vect{X}} \cdot \vect{P^{cm}} + R(\vect{X},0)(\vect{B^{bd}}+ \vect{B^{act}}).
    \label{eq:sm_lcmv2}
\end{equation}

\textbf{Weak Form of the Conservation of Momentum in Both Views.} The weak form of an equation means that we can derive its weak form from the equation but can't derive the equation from its weak form. The weak forms of the conservation of momentum in both views are extremely important to derive our discretized Crowd MPM. We start with the Lagrangian view. We ignore the body force and the active force for simplicity:
\begin{equation}
    R(\vect{X},0)\vect{A}(\vect{X},t) = \nabla^{\vect{X}} \cdot \vect{P^{cm}}.
    \label{eq:sm_wfl_rap}
\end{equation}
Then, we have:
\begin{align}
    R_0A_i= \sum_j \frac{\partial P_{ij}^{cm}}{\partial X_j} = \sum_j P_{ij,j}^{cm} ,
\end{align}
where $R_0=R(\vect{X},0)$, $A_i$ is the ith component of $\vect{A}(\vect{X},t)$, $P_{ij}^{cm}$ is the element at the ith row and jth column in $\vect{P^{cm}}$. For efficient expression, the summation is implied on the repeated index. Therefore, we have:
\begin{align}
    R_0A_i = P_{ij,j}^{cm}.
\end{align}
Then, we derive the weak form of \cref{eq:sm_wfl_rap} by computing the dot product between an arbitrary function $\vect{Q}(\cdot,t):\Omega^0 \rightarrow \mathbb{R}^d$ and two sides of \cref{eq:sm_wfl_rap}, respectively, and integrate over $\Omega^0$:
\begin{align}
    &\int_{\Omega^0} Q_i(\vect{X}, t)R(\vect{X}, 0)A_i(\vect{X}, t) d\vect{X} = \notag \\
    & \int_{\Omega^0} Q_i(\vect{X}, t)P_{ij,j}^{cm}(\vect{X}, t) dX = \notag \\ 
    & \int_{\Omega^0}((Q_i(\vect{X}, t)P_{ij}^{cm}(\vect{X}, t))_{,j}-Q_{i,j}(\vect{X}, t)P_{ij}^{cm}(\vect{X}, t))d\vect{X} \notag = \\
    & \int_{\partial\Omega^0}Q_i P_{ij}^{cm} N_j ds(\vect{X}) - \int_{\Omega^0}Q_{i,j} P_{ij}^{cm} d\vect{X}.
\end{align}
The $P_{ij}^{cm} N_j$ would be determined by a boundary condition. Assuming that $\vect{T}(\vect{X},t)$ is the boundary force per unit reference area, we have $T_i=P_{ij}^{cm} N_j$. As a result, we have that for an arbitrary $\vect{Q}(\cdot,t):\Omega^0 \rightarrow \mathbb{R}^d$:
\begin{align}
 \int_{\Omega^0} Q_i R_0 A_i d\vect{X} =  \int_{\partial \Omega^0} Q_i T_i ds(\vect{X}) - \int_{\Omega^0} Q_{i, j}P_{ij}^{cm}dX.
\end{align}
Considering the full \cref{eq:sm_lcmv2}, \ie \cref{eq:sm_lcmv}, we have the final weak form of the conservation of momentum in the Lagrangian view:
\begin{align}
 \int_{\Omega^0} Q_i R_0 A_i d\vect{X} =  \int_{\partial \Omega^0} Q_i T_i ds(\vect{X}) - \int_{\Omega^0} Q_{i, j}P_{ij}^{cm}d\vect{X} \notag \\
 + \int_{\Omega^0} Q_i R_0 B^{bd}_i d\vect{X} + \int_{\Omega^0} Q_i R_0 B^{act}_i d\vect{X}.
 \label{eq:sm_wf_l}
\end{align}
Then we push forward \cref{eq:sm_wf_l} to get the weak form of the conservation of momentum in the Eulerian view. The push forward of the left side of \cref{eq:sm_wf_l} is derived by introducing $\vect{q}$ that is the push forward of $\vect{Q}$:
\begin{align}
 \int_{\Omega^0} Q_i R_0 A_i d\vect{X} =  \int_{\Omega^t} q_i(\vect{x}, t) \rho(\vect{x}, t) a_i(\vect{x}, t) d\vect{x}.
 \label{eq:sm_wfe_d1}
\end{align}
Similarly, let $\vect{t}$ be the push forward of $\vect{T}$, we have:
\begin{align}
 \int_{\partial \Omega^0} Q_i T_i ds(\vect{X}) = \int_{\partial \Omega^t} q_i(\vect{x}, t) t_i(\vect{x}, t) ds(\vect{x}).  
  \label{eq:sm_wfe_d2}
\end{align}
Before giving the push forward of the second integral on the right side of \cref{eq:sm_wf_l}, we analyze $Q_{i,j}$:
\begin{equation}
Q_{i, j} = \frac{\partial Q_i}{\partial X_j} = \frac{\partial q_i}{\partial x_k} \frac{\partial x_k}{ \partial X_j} = q_{i,k}F_{kj},
\label{eq:sm_wf_e_r2d1}
\end{equation}
where the summation is implied on the repeated index $k$, $F_{kj}$ is the element in $\vect{F}$ defined by \cref{{eq:defMt}}, and $\vect{x} = \Phi(\vect{X}, t)$. Recall that $\vect{P^{cm}}=J\vect{\sigma^{cm}}\vect{F}^{-T}$, then we have $\vect{\sigma^{cm}}=\frac{1}{J}\vect{P^{cm}}\vect{F}^{T}$. Therefore, we the relationship between $\vect{\sigma^{cm}}$ and $\vect{P^{cm}}$:
\begin{align}
    \sigma^{cm}_{ik}==\frac{1}{J}P^{cm}_{ij}F_{kj}.
    \label{eq:sm_wf_e_r2d2}
\end{align}
Combining \cref{eq:sm_wf_e_r2d1} and \cref{eq:sm_wf_e_r2d2} let us derive the push forward of the second integral on the right side of \cref{eq:sm_wf_l}:
\begin{align}
    \int_{\Omega^0} Q_{i, j}P_{ij}^{cm}d\vect{X} = \int_{\Omega^t} q_{i, k}(\vect{x}, t)\sigma_{ik}^{cm}(\vect{x}, t)d\vect{x}. 
    \label{eq:sm_wfe_d3}
\end{align}
As $\vect{b}^{bd}$ and $\vect{b}^{act}$ are the natural push forward of $\vect{B}^{bd}$ and $\vect{B}^{act}$, it is easy to derive the push forward of the remaining integrals in \cref{eq:sm_wf_l}:
\begin{align}
     &\int_{\Omega^0} Q_i R_0 B^{bd}_i d\vect{X} =
      \int_{\Omega^t} q_i(\vect{x}, t) \rho(\vect{x}, t) b^{bd}_i(\vect{x}, t) d\vect{x} \notag \\
     &\int_{\Omega^0} Q_i R_0 B^{act}_i d\vect{X} =       \int_{\Omega^t} q_i(\vect{x}, t) \rho(\vect{x}, t) b^{act}_i(\vect{x}, t) d\vect{x}.
     \label{eq:sm_wfe_d4}
\end{align}
Combing \cref{eq:sm_wfe_d1}, \cref{eq:sm_wfe_d2}, \cref{eq:sm_wfe_d3}, and \cref{eq:sm_wfe_d4} results in the final weak form of the conservation of momentum in the Eulerian view: 
\begin{align}
    \int_{\Omega^t} q_i \rho a_i d\vect{x} = \int_{\partial \Omega^t} q_i t_i ds(\vect{x}) - \int_{\Omega^t} q_{i, k} \sigma_{ik}^{cm}d\vect{x} \notag \\
    +\int_{\Omega^t} q_i \rho b^{bd}_i d\vect{x} + \int_{\Omega^t} q_i \rho b^{act}_i d\vect{x}.
\end{align}

\end{document}